\pdfoutput=1

\documentclass[11pt, letterpaper, twoside]{article}

\usepackage{emnlp2021}

\usepackage{times}
\usepackage{latexsym}

\usepackage[T1]{fontenc}

\usepackage[utf8]{inputenc}

\usepackage{microtype}

\usepackage{graphicx}
\usepackage{subfigure}
\usepackage{multirow}

\usepackage[normalem]{ulem}
\useunder{\uline}{\ul}{}

\usepackage{dblfloatfix}

\title{Navigating the Kaleidoscope of COVID-19 Misinformation Using Deep Learning}

\author{Yuanzhi Chen \\
  University of Nebraska-Lincoln\\
  NE 68588, USA \\
  \texttt{yuanzhi@huskers.unl.edu} \\\And
  Mohammad Rashedul Hasan \\
  University of Nebraska-Lincoln\\
  NE 68588, USA \\
   \texttt{hasan@unl.edu} \\}

\begin{document}
\maketitle
\begin{abstract}


Irrespective of the success of the deep learning-based mixed-domain transfer learning approach for solving various Natural Language Processing tasks, it does not lend a generalizable solution for detecting misinformation from COVID-19 social media data. Due to the inherent complexity of this type of data, caused by its dynamic (context evolves rapidly), nuanced (misinformation types are often ambiguous), and diverse (skewed, fine-grained, and overlapping categories) nature, it is imperative for an effective model to capture both the local and global context of the target domain. By conducting a systematic investigation, we show that: (i) the deep Transformer-based pre-trained models, utilized via the mixed-domain transfer learning, are only good at capturing the local context, thus exhibits poor generalization, and (ii) a combination of shallow network-based domain-specific models and convolutional neural networks can efficiently extract local as well as global context directly from the target data in a hierarchical fashion, enabling it to offer a more generalizable solution.

\end{abstract}

\section{Introduction}
\label{intro}

Since the start of the Coronavirus or COVID-19 pandemic, online social media (e.g., Twitter) has become a conduit for rapid propagation of misinformation \cite{johnson2020covid19}. Although misinformation is considered to be created without the intention of causing harm \cite{Lazer1094}, it can wreak havoc on society \cite{Ciampagliapub.1093084646, neuman_2020,hamilton_2020} and disrupt democratic institutions \cite{Ciampaglia_Mantzarlis_Maus_Menczer_2018}. Misinformation in general, and COVID-19 misinformation in particular, has become a grave concern for the policymakers due to its fast propagation via online social media. A recent study shows that the majority of the COVID-19 social media data is rife with misinformation \cite{brennen_simon_howard_nielsen_2020}. The first step towards preventing misinformation is to \textbf{detect misinformation} in a timely fashion.

Building automated systems for misinformation detection from social media data is a Natural Language Processing (NLP) task. Various deep learning models have been successfully employed for this type of NLP task of text classification \cite{Kim:2014,Conneau:2017, Wang:2017, Tai:2015, Zhou:2016}. These models learn language representations from a domain, which are then used as numeric features in supervised classification. Due to the prohibitive cost of acquiring labeled data on COVID-19 misinformation, training deep learning models directly using the target data is not a suitable approach.


\paragraph{Background.} An alternative approach for detecting COVID-19 misinformation from \textbf{small labeled data} is transfer learning \cite{hossain-etal-2020-covidlies}. The dominant paradigm of transfer learning employs a \textbf{mixed-domain} strategy in which representations learned from a general domain (source data) by using domain-agnostic models are transferred into a specific domain (target data) \cite{Pan:2009}. Specifically, it involves creating a pre-trained model (PTM) that learns embedded representations from general-purpose unlabeled data, then adapting the model for a downstream task using the labeled target data \cite{Minaee:2021, Qiu:2020}. 

Two types of neural networks can be used to create PTMs, i.e., shallow and deep. The shallow models such as Word2Vec \cite{mikolov2013efficient} and GloVe \cite{pennington-etal-2014-glove} learn word embeddings that capture semantic, syntactic, and some global relationships \cite{Levy:2014, Srinivasan:2019} of the words from the source text using their co-occurrence information. However, these PTMs do not capture the context of the text \cite{Qiu:2020}. On the other hand, deep PTMs can learn contextual embeddings, i.e., language models \cite{Yoav:2017}. 

Two main approaches for creating deep PTMs are based on sequential and non-sequential models. The sequential Recurrent Neural Network \cite{Pengfei:2016} based model such as ELMo (Embeddings from Language Models) \cite{peters2018deep} is equipped with long short-term memory to capture the local context of a word in sequential order. The non-sequential Transformer \cite{Vaswani:2017} based models such as OpenAI GPT (Generative Pre-training) \cite{Radford:2018}, BERT (Bidirectional Encoder Representation from Transformer) \cite{devlin2019bert}, and XLNet \cite{yang2020xlnet} utilize the attention mechanism \cite{Bahdanau:2016} for learning universal language representation from general-purpose very large text corpora such as Wikipedia and BookCorpus \cite{devlin2019bert} as well as from web crawls \cite{liu2019roberta}. While GPT is an autoregressive model that learns embeddings by predicting words based on previous predictions, BERT utilizes the autoencoding technique based on bi-directional context modeling \cite{Minaee:2021}. XLNet leverages the strengths of autoregressive and autoencoding PLMs \cite{yang2020xlnet}. 

Unlike the Transformer-based deep PTMs, the shallow Word2Vec and GloVe as well as the deep ELMo PTMs are used only as feature extractors. These features are fed into another model for the downstream task of classification, which needs to be trained from scratch using the target data. The deep PTM based mixed-domain transfer learning has achieved state-of-the-art (SOTA) performance in many NLP tasks including text classification \cite{Minaee:2021}.  

Irrespective of the success of the mixed-domain SOTA transfer learning approach for text classification, there has been no study to understand how effective this approach is for navigating through the kaleidoscope of COVID-19 misinformation. Unlike the curated static datasets on which this approach is tested \cite{Minaee:2021}, the dynamic landscape of the COVID-19 social media data has not been fully explored. Some key properties of the COVID-19 data hitherto identified are: (i) The COVID-19 misinformation spreads faster on social media than any other form of health misinformation \cite{johnson2020covid19}. As a consequence, the misinformation narrative evolves rapidly \cite{cui2020coaid}. (ii) The COVID-19 misinformation categories are heavily-skewed \cite{cui2020coaid, memon2020characterizing} and fine-grained \cite{memon2020characterizing}. (iii) The COVID-19 social media misinformation types are often ambiguous (e.g., fabricated, reconfigured, satire, parody)  \cite{brennen_simon_howard_nielsen_2020} and categories may not be mutually exclusive \cite{memon2020characterizing}. These properties pose a unique challenge for the mixed-domain SOTA transfer learning approach for creating an effective solution to the COVID-19 misinformation detection problem. 

Previously, it has been shown that the transfer learning approach generalizes poorly when the domain of the source dataset is significantly different from that of the target dataset \cite{peters2019tune}. On the other hand, \textbf{domain-specific models (DSM)}, which learn representations from domains that are similar to the target domain, provide a generalizable solution for the downstream NLP task \cite{beltagy2019scibert, Lee:2019, Gu:2021}. These models are better at capturing the context of the target domain. However, the efficacy of the DSM-based approach for addressing the COVID-19 misinformation detection problem has not also been investigated. 

\vspace{3.00mm}
\noindent \textbf{In this paper}, we conduct a systematic extensive study to understand the \textbf{scope and limitations} of the mixed-domain transfer learning approach as well as the DSM-based approach to detect COVID-19 misinformation on social media. We use both shallow and deep PTMs for the mixed-domain transfer learning experimentations. The deep PTMs include BERT, XLNet, and two variants of BERT, i.e., RoBERTa \cite{liu2019roberta} and ALBERT \cite{lan2020albert}. While these attention mechanism-based Transformer models are good at learning contextual representations, their ability to learn \textbf{global relationships} among the words in the source text is limited \cite{lu2020vgcnbert}.

The DSMs used in our study are based on shallow architectures. We argue that shallow architectures can be trained efficiently using the limited available domain data. Specifically, we pre-train the DSMs using the small social media data on COVID-19. The shallow DSM-based approach is examined in \textbf{two dimensions: graph-based DSM and non-graph DSM}. The graph-based Text GCN \cite{yao2018graph} model can explicitly capture the \textbf{global relationships} (from term co-occurrence) by leveraging the graph structure of the text. It creates a heterogeneous word document graph with words and documents as nodes for the whole corpus, and turns document classification problem into a node classification problem. We include another graph-based model in our study, i.e., the VGCN-BERT \cite{lu2020vgcnbert}. It combines the strength of Text GCN (to capture global relationships) and BERT (to capture local relationships).

The non-graph DSM models such as Word2Vec and GloVe can mainly capture \textbf{local relationships} among the words of the source text, represented in the latent space of their word embeddings. For extracting \textbf{global relationships} from these embeddings, we utilize a Convolutional Neural Network (CNN). Specifically, we use the word embeddings as input features to a CNN with a one-dimensional kernel \cite{Kim:2014}, which then learns global relationships as high-level features. We \textbf{hypothesize} that the local and global relationships should improve the generalization capability of the non-graph DSM+CNN approach.



We evaluate the \textbf{generalizability} of the above-mentioned diverse array of NLP techniques via a set of studies that explore \textbf{various dimensions of the COVID-19 data}. We focus on the Twitter social media platform because of its highest number of news-focused users \cite{hughes_wojcik_2020}. In addition to analyzing the tweet messages, we use online news articles referred to in the tweets. Our study spans along multiple dimensions of the COVID-19 data that include temporal dimension (the context in the dataset evolves), length dimension (short text such as tweets vs. lengthy text such as news articles), size dimension (small dataset vs. large dataset), and classification-level dimension (binary vs. multi-class data).

\paragraph{Contributions.} We design a novel study for examining the generalizability of a diverse set of deep learning NLP techniques on the multi-dimensional space of COVID-19 online misinformation landscape. Our main contributions are as follows.

\begin{itemize}

  \item We identify the unique challenges for the deep learning based NLP techniques to detect misinformation from COVID-19 social media data. 
  
  \item We argue that an effective model for this type of data must capture both the local and the global context of the domain in its latent space of embeddings.
    
  \item We show that the mixed-domain deep learning SOTA transfer learning approach is not always effective. 
  
  \item We find that the shallow CNN classifier initialized with word embeddings learned via the non-graph DSMs is more effective across most of the dimensions of the COVID-19 data space, especially when the labeled target data is small.
  
  \item We \textbf{explain} why the Transformer-based mixed-domain transfer learning approach is not effective on COVID-19 data as well as why the non-graph DSM+CNN may offer a more generalizable solution.

\end{itemize}

The rest of the paper is organized as follows. In section 2, we present the diverse NLP techniques, analyze the multi-dimensional datasets, and describe the study design. Results obtained from the experiments are provided in section 3 followed by a detailed analysis. Section 4 presents the conclusion. Appendix provides related work, additional analysis of the datasets, and experiment setting.


\section{Method}

First, we describe how we obtained various PTMs and created DSM embeddings for different models as well as how we fine-tuned/trained the classifiers for the studies. Then, we discuss the datasets and the study design. 

\subsection{Mixed-Domain Transfer Learning}
\label{tlm}


We use the following PTMs: BERT, RoBERTa, ALBERT, XLNet, ELMo, Word2Vec, and GloVe. 

\textbf{Deep PTMs: }We get the BERT base model (uncased) for sequence classification from the Hugging Face library \cite{wolf-etal-2020-transformers}. The embedding vectors are 768-dimensional. This BERT PTM adds a single linear layer on top of the BERT base model. The pretrained weights of all hidden layers of the PTM and the randomly initialized weights of the top classification layer are adapted during fine-tuning using a target dataset. The XLNet is obtained from the Hugging Face library \cite{wolf-etal-2020-transformers} and fine-tuned similar to BERT. Its embedding vectors are 768-dimensional. The RoBERTa (obtained from \cite{wolf-etal-2020-transformers}) and ALBERT (obtained from  \cite{maiya2020ktrain}) are used by first extracting embeddings from their final layer and then adding linear layers. While the RoBERTa embeddings are 768-dimensional, the ALBERT embeddings are 128-dimensional.

\textbf{Shallow PTMs: }We get the ELMo embeddings from TensorFlow-Hub \cite{tensorflow2015-whitepaper}. Each embedding vector has a length of 1024. The Word2Vec embeddings are obtained from Google Code \cite{google}. The embedding vectors are 300-dimensional. We get the GloVe pretrained 300-dimensional embeddings from \cite{pennington-etal-2014-glove}.

\textbf{CNN:} The ELMo, Word2Vec, and GloVe embeddings are used to train a CNN classifier with a single hidden layer \cite{Kim:2014}. The first layer is the embedding layer. Its dimension varies based on the dimension of pretrained embeddings. The second layer is the one-dimensional convolution layer that consists of 100 filters of dimension 5 x 5 with ``same'' padding and ReLU activation. The third layer is a one-dimensional global max-pooling layer, and the fourth layer is a dense layer with 100 units along with ReLU activation. The last layer is the classification layer with softmax activation. We use this setting for the CNN architecture as it was found empirically optimal in our experiments. We use cross-entropy as loss function, Adam as the optimizer, and a batch size of 128. The embedding vectors are kept fixed during the training \cite{Kim:2014}.


\subsection{Domain-Specific Model (DSM) based Learning}

We create the DSMs using two approaches: graph-based and non-graph. For the graph-based approach, we use the following models: Text GCN and VGCN-BERT. For training the Text GCN model, we pre-process the data as follows. First, we clean the text by removing stop words and rare words whose frequencies are less than 5. Then, we build training, validation, and test graphs using the cleaned text. Finally, we train the GCN model using training and validation graphs and test the model using a test graph. During the training, early stopping is used.

For training the VGCN-BERT model, first, we clean the data that includes removing spaces, the special symbols as well as URLs. Then, the BERT tokenizer is used to create BERT vocabulary from the cleaned text. The next step is to create training, validation, and the test graphs. The last step is training the VGCN-BERT model. During the training, the model constructs embeddings from word and vocabulary GCN graph.

For the non-graph approach, we create embeddings from the target dataset by using the Word2Vec and GloVe models. First, we pre-process the raw text data by converting the text (i.e., a list of sentences) into a list of lists containing tokenized words. During tokenization, we convert words to lowercase, remove words that are only one character, and lemmatize the words. We add bigrams that appear 10 times or more to our tokenized text. The bigrams allow us to create phrases that could be helpful for the model to learn and produce more meaningful representations. Then, we feed our final version of the tokenized text to the Word2Vec and the GloVe model for creating embeddings. After we obtain the embeddings, we use them to train the CNN classifier described in the previous sub-section, except that the domain-specific word embeddings are adapted during the training.


\subsection{Dataset}

We use two COVID-19 datasets for the study, i.e., CoAID \cite{cui2020coaid} and CMU-MisCov19 \cite{memon2020characterizing}. 

The CoAID dataset contains two types of data: true information and misinformation. We use this dataset to investigate the generalizability of the models along three dimensions.

\begin{itemize}

    \item \textbf{Temporal dimension}: Train a model using data from an earlier time, then test its generalizability at different times in the future.
    
    \item \textbf{Size dimension}: Train models by varying the size of the training dataset.
    
    \item \textbf{Length dimension}: Train models by varying the length of the samples, e.g., tweet (short-length data) and news articles (lengthy data).  
    
    
\end{itemize}

The CMU-MisCov19 dataset is used to analyze a model's performance in fine-grained classification.


\subsubsection{CoAID: Binary Classification}

%
%

The CoAID dataset \cite{cui2020coaid} is used for binary classification since it has only two labels: 0 for misinformation and 1 for true information. This dataset contains two types of data: online news articles on COVID-19 and tweets related to those articles. Datasets of these two categories were collected at four different months in 2020: May, July, September, and November. Thus, the total number of CoAID datasets is 8. The class distribution is \textbf{heavily skewed} with significantly more true information samples than misinformation samples. Sample distribution per class (both for the tweets and news articles) is given in the appendix.




\subsubsection{CMU-MisCov19: Fine-Grained Classification}

The CMU-MisCov19 dataset contains 4,573 annotated tweets \cite{memon2020characterizing}. The tweets were collected on three days in 2020: March 29, June 15, and June 24. The categories are fine-grained comprising of 17 classes with \textbf{skewed distribution}. This dataset does not have any true information category. Its sample distribution per class is given in the appendix.

\subsection{Context Evolution in the COVID-19 Social Media Data}

We use the CoAID dataset to understand whether the context of the COVID-19 text evolves. To detect a change in the context over time, we investigate how the distribution of the high-frequency terms evolve for the two categories of the data: tweets and news articles. For each category, we select the top 10 high-frequency words from the 4 non-overlapping datasets belonging to 4 subsequent months, i.e., May, July, September, and November in 2020. Our goal is to determine whether there exists a temporal change in the distribution of high-frequency words.

\begin{figure}[!htb]
    \begin{center}
        \subfigure[May]{\label{tweet-may}\includegraphics[width=2.80in]{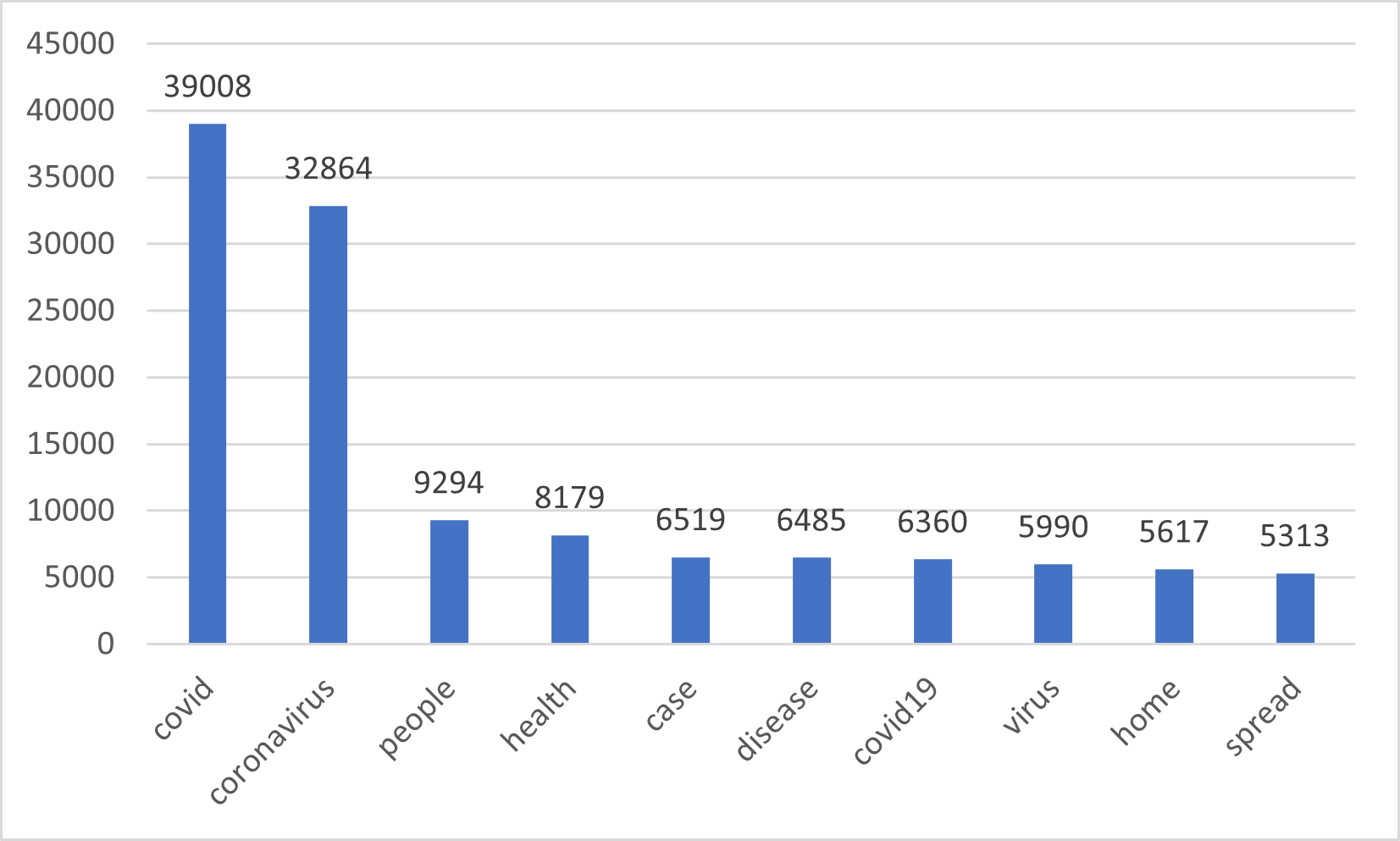}} \quad \vspace{-3.00mm}	
        \subfigure[July]{\label{tweet-july}\includegraphics[width=2.80in]{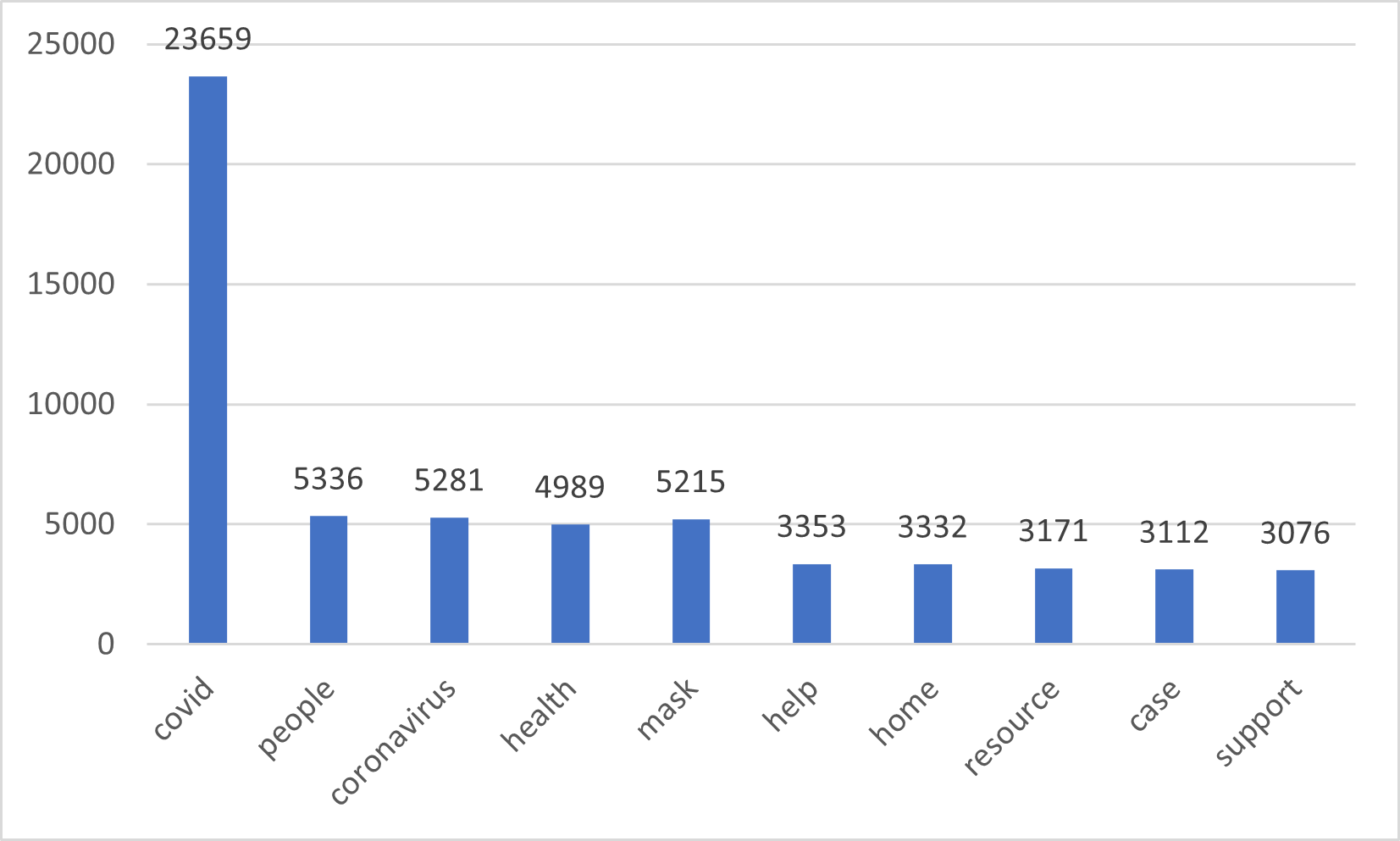}} \\	 
        \subfigure[September]{\label{tweet-sept}\includegraphics[width=2.80in]{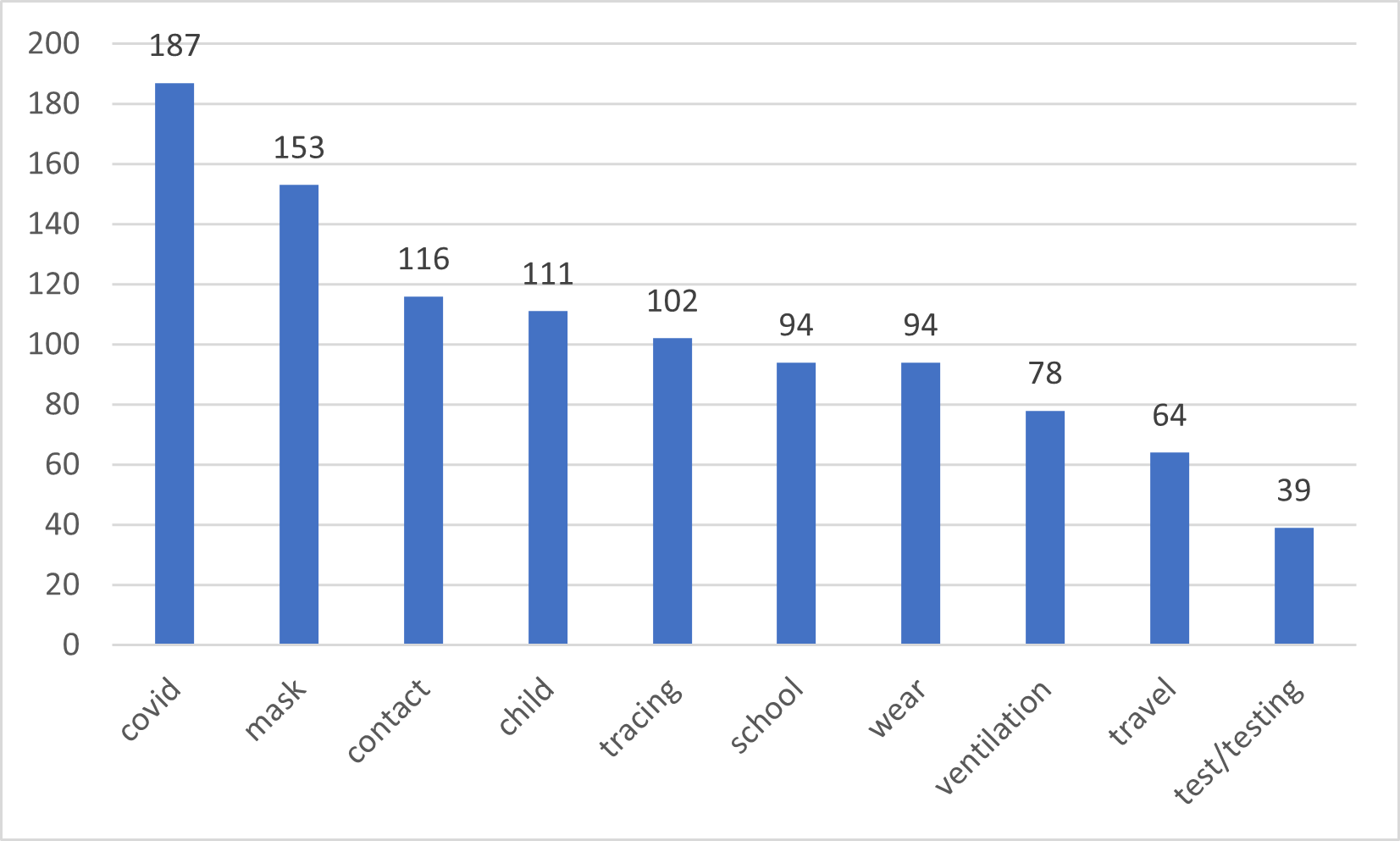}} \quad \vspace{-3.00mm}
        \subfigure[November]{\label{tweet-nov}\includegraphics[width=2.80in]{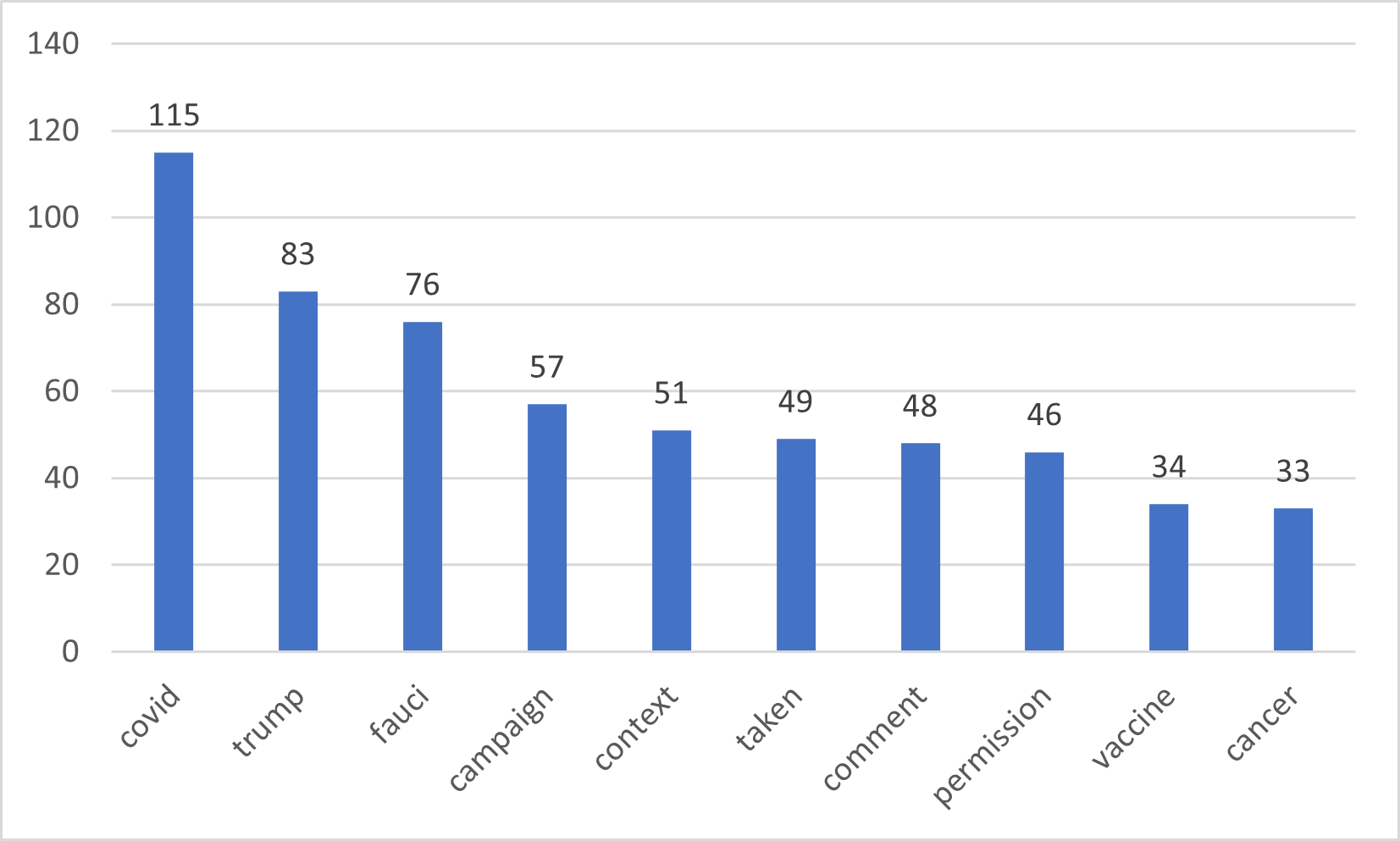}} \quad

        \caption{CoAID (Tweets): Frequency of top 10 words in four datasets from four subsequent months.}
	 \vspace{-5.00mm}			
        \label{tweet-top10}
    \end{center}
\end{figure}

Figure \ref{tweet-top10} shows context evolution in the tweets category. We see that during May, the two high-frequency words were covid and coronavirus. The frequent words represent broader concepts such as health, disease, spread, etc. However, over time the context shifted towards more loaded terms. For example, in July two new high-frequency words, such as mask and support, emerged. Then, in September words like contact, school, child, and travel became prominent. Finally, during November, we observe \textbf{a sharp change} in the nature of the frequent words. Terms with strong political connotations (e.g., trump, fauci, campaign, and vaccine) started emerging. The evolution in the high-frequency words indicates a temporal shift in the context in the tweets dataset. We observe similar context evolution in the news articles dataset, reported in the Appendix with additional analysis.



\subsection{Study Design}

We describe the design of the studies for comparing the NLP approaches for misinformation detection.


\subsubsection{Study 1}

Study 1 is designed to explore a model's generalizability in the \textbf{temporal dimension} of the data. We fine-tune/train a model using CoAID data collected from May 2020 and test it using data obtained from 3 different months in ``future'': July, September, and November. The following models are tested in this study: BERT (Mixed-domain Transfer Learning), ELMo (Mixed-domain Transfer Learning), Word2Vec (Mixed-domain Transfer Learning and DSM-based), GloVe (Mixed-domain Transfer Learning and DSM-based), Text-GCN (DSM-based), and VGCN-BERT (DSM-based).

\subsubsection{Study 2}

Study 2 is designed to test the performance of a model along the \textbf{length dimension}  of the data. We use both short-length data (tweets) and lengthy data (news articles). Specifically, we train a model using the CoAID Twitter dataset to understand a model's performance on the short-length data. Then, we train a model using the CoAID news articles dataset to study a model's performance on the lengthy data. The models used in this study are the same as in Study 1.

\subsubsection{Study 3}

In study 3, we evaluate a model along the \textbf{size dimension}  of the data. We replicate studies 1 and 2 using a large target dataset, which is created by merging the datasets from May, July, and September. The November dataset is used as the test set. We experiment with two models for this study: BERT (Mixed-Domain Transfer Learning) and Word2Vec (DSM-based).

\subsubsection{Study 4}

To further study the effectiveness of the Transformer-based mixed-domain transfer learning approach, we experiment with two variants of the BERT PTM, i.e., RoBERTa and ALBERT. In addition to this, we study the performance of an autoregressive model XLNet that induces the strength of BERT. For this study, we only use the news articles dataset.


\subsubsection{Study 5}

Study 5 is designed to test a model's performance on the fined-grained CMU-MisCov19 dataset. The models tested are the same as in Study 1.

%

\section{Results and Analysis}

\begin{table*}[!htb]

\begin{center}
\scalebox{0.6}{
\begin{tabular}{|l|c|l|l|l|l|l|c|l|l|l|l|l|c|l|l|l|l|l|}
\hline
\multicolumn{1}{|c|}{\textbf{Train: May}} &
  \multicolumn{6}{c|}{\textbf{Test: July}} &
  \multicolumn{6}{c|}{\textbf{Test: September}} &
  \multicolumn{6}{c|}{\textbf{Test: November}} \\ \hline
\textbf{Model} &
  \multicolumn{3}{c|}{\textbf{True Information}} &
  \multicolumn{3}{c|}{\textbf{Misinformation}} &
  \multicolumn{3}{c|}{\textbf{True Information}} &
  \multicolumn{3}{c|}{\textbf{Misinformation}} &
  \multicolumn{3}{c|}{\textbf{True Information}} &
  \multicolumn{3}{c|}{\textbf{Misinformation}} \\ \hline
 &
  \multicolumn{1}{l|}{P} &
  R &
  F1 &
  P &
  R &
  F1 &
  \multicolumn{1}{l|}{P} &
  R &
  F1 &
  P &
  R &
  F1 &
  \multicolumn{1}{l|}{P} &
  R &
  F1 &
  P &
  R &
  F1 \\ \hline
\multicolumn{19}{|c|}{\textbf{Mixed-Domain Transfer Learning}} \\ \hline
 &
  \multicolumn{6}{c|}{Accuracy = 0.978} &
  \multicolumn{6}{c|}{Accuracy = 0.942} &
  \multicolumn{6}{c|}{Accuracy = 0.990} \\ \cline{2-19} 
\multirow{-2}{*}{\textbf{BERT}} &
  \multicolumn{1}{l|}{0.981} &
  0.997 &
  0.989 &
  0.281 &
  {\color[HTML]{000000} 0.079} &
  0.116 &
  \multicolumn{1}{l|}{0.946} &
  0.996 &
  0.970 &
  0.00 &
  0.00 &
  0.00 &
  \multicolumn{1}{l|}{1.00} &
  0.990 &
  0.995 &
  0.500 &
  1.00 &
  0.670 \\ \hline
 &
  \multicolumn{6}{c|}{Accuracy = 0.979} &
  \multicolumn{6}{c|}{Accuracy = 0.941} &
  \multicolumn{6}{c|}{Accuracy = 0.990} \\ \cline{2-19} 
\multirow{-2}{*}{\textbf{ELMo}} &
  \multicolumn{1}{l|}{0.979} &
  1.00 &
  0.989 &
  0.00 &
  0.00 &
  0.00 &
  \multicolumn{1}{l|}{0.945} &
  0.995 &
  0.970 &
  0.00 &
  0.00 &
  0.00 &
  \multicolumn{1}{l|}{0.990} &
  1.00 &
  0.995 &
  0.00 &
  0.00 &
  0.00 \\ \hline
 &
  \multicolumn{6}{c|}{Accuracy = 0.979} &
  \multicolumn{6}{c|}{Accuracy = 0.932} &
  \multicolumn{6}{c|}{Accuracy = 0.969} \\ \cline{2-19} 
\multirow{-2}{*}{\textbf{Word2Vec}} &
  \multicolumn{1}{l|}{0.979} &
  1.00 &
  0.989 &
  0.00 &
  0.00 &
  0.00 &
  \multicolumn{1}{l|}{0.949} &
  0.980 &
  0.964 &
  0.20 &
  0.09 &
  0.12 &
  \multicolumn{1}{l|}{0.99} &
  0.979 &
  0.984 &
  0.00 &
  0.00 &
  0.00 \\ \hline
 &
  \multicolumn{6}{c|}{Accuracy = 0.979} &
  \multicolumn{6}{c|}{Accuracy = 0.943} &
  \multicolumn{6}{c|}{Accuracy = 1.00} \\ \cline{2-19} 

\multirow{-2}{*}{\color[HTML]{FE0000} {\ul \textbf{GloVe}}}&

  \multicolumn{1}{l|}{0.979} &
  0.999 &
  0.989 &
  0.31 &
  0.02 &
  0.03 &
  \multicolumn{1}{l|}{0.946} &
  0.998 &
  0.971 &
  0.00 &
  0.00 &
  0.00 &
  \multicolumn{1}{l|}{1.00} &
  1.00 &
  1.00 &
  {\color[HTML]{FE0000} {\ul \textbf{1.00}}} &
  {\color[HTML]{FE0000} {\ul \textbf{1.00}}} &
   {\color[HTML]{FE0000} {\ul \textbf{1.00}}}  \\ \hline
\multicolumn{19}{|c|}{\textbf{DSMs: Graph-based}} \\ \hline
 &
  \multicolumn{6}{c|}{Accuracy = 0.979} &
  \multicolumn{6}{c|}{Accuracy = 0.946} &
  \multicolumn{6}{c|}{Accuracy = 0.99} \\ \cline{2-19} 
\multirow{-2}{*}{\textbf{Text GCN}} &
  \multicolumn{1}{l|}{0.979} &
  1.00 &
  0.989 &
  0.029 &
  0.00 &
  0.00 &
  \multicolumn{1}{l|}{0.00} &
  0.00 &
  0.00 &
  0.95 &
  1.00 &
  0.97 &
  \multicolumn{1}{l|}{0.00} &
  0.00 &
  0.00 &
  0.99 &
  1.00 &
  0.99 \\ \hline
 &
  \multicolumn{6}{c|}{Accuracy = 0.978} &
  \multicolumn{6}{c|}{Accuracy = 0.946} &
  \multicolumn{6}{c|}{Accuracy = 0.971} \\ \cline{2-19} 

\multirow{-2}{*}{\color[HTML]{FE0000} {\ul \textbf{VGCN-BERT}}}&

  \multicolumn{1}{l|}{0.979} &
  0.999 &
  0.989 &
  {\color[HTML]{FE0000} {\ul \textbf{0.495}}} &
  0.019 &
  0.035 &
  \multicolumn{1}{l|}{0.947} &
  0.999 &
  0.972 &
  {\color[HTML]{FE0000} {\ul \textbf{0.287}}} &
  0.029 &
  0.052 &
  \multicolumn{1}{l|}{0.994} &
  0.977 &
  0.984 &
  0.21 &
  0.389 &
  0.246 \\ \hline
\multicolumn{19}{|c|}{\textbf{DSMs: Non-Graph + CNN}} \\ \hline
 &
  \multicolumn{6}{c|}{Accuracy = 0.977} &
  \multicolumn{6}{c|}{Accuracy = 0.946} &
  \multicolumn{6}{c|}{Accuracy = 0.99} \\ \cline{2-19} 
\multirow{-2}{*}{\textbf{Word2Vec}} &
  \multicolumn{1}{l|}{0.979} &
  0.997 &
  0.988 &
  0.11 &
  0.02 &
  0.03 &
  \multicolumn{1}{l|}{0.946} &
  1.00 &
  0.972 &
  0.00 &
  0.00 &
  0.00 &
  \multicolumn{1}{l|}{0.99} &
  1.00 &
  0.995 &
  0.00 &
  0.00 &
  0.00 \\ \hline
 &
  \multicolumn{6}{c|}{Accuracy = 0.978} &
  \multicolumn{6}{c|}{Accuracy = 0.946} &
  \multicolumn{6}{c|}{Accuracy = 0.979} \\ \cline{2-19} 
\multirow{-2}{*}{\textbf{GloVe}} &
  \multicolumn{1}{l|}{0.979} &
  0.999 &
  0.989 &
  0.23 &
  0.02 &
  0.03 &
  \multicolumn{1}{l|}{0.946} &
  1.00 &
  0.972 &
  0.00 &
  0.00 &
  0.00 &
  \multicolumn{1}{l|}{0.99} &
  0.99 &
  0.99 &
  0.00 &
  0.00 &
  0.00 \\ \hline
\end{tabular}}
\vspace{-2.00mm}	
\caption{Study 1 \& 2: CoAID - Tweet (Temporal \& Text Length Dimension). Best results, as well as the optimal models, are highlighted in red. \textbf{None of the models generalize well on the tweet data.}}
\label{CoAID-Tweet}
\end{center}
\vspace{-2.00mm}

\end{table*}

\begin{table*}[!htb]

\begin{center}
\scalebox{0.6}{
\begin{tabular}{|l|c|l|l|l|l|l|c|l|l|l|l|l|c|l|l|l|l|l|}
\hline
\multicolumn{1}{|c|}{\textbf{Train: May}} &
  \multicolumn{6}{c|}{\textbf{Test: July}} &
  \multicolumn{6}{c|}{\textbf{Test: September}} &
  \multicolumn{6}{c|}{\textbf{Test: November}} \\ \hline
\textbf{Model} &
  \multicolumn{3}{c|}{\textbf{True Information}} &
  \multicolumn{3}{c|}{\textbf{Misinformation}} &
  \multicolumn{3}{c|}{\textbf{True Information}} &
  \multicolumn{3}{c|}{\textbf{Misinformation}} &
  \multicolumn{3}{c|}{\textbf{True Information}} &
  \multicolumn{3}{c|}{\textbf{Misinformation}} \\ \hline
 &
  \multicolumn{1}{l|}{P} &
  R &
  F1 &
  P &
  R &
  F1 &
  \multicolumn{1}{l|}{P} &
  R &
  F1 &
  P &
  R &
  F1 &
  \multicolumn{1}{l|}{P} &
  R &
  F1 &
  P &
  R &
  F1 \\ \hline
\multicolumn{19}{|c|}{\textbf{Mixed-Domain Transfer Learning}} \\ \hline
 &
  \multicolumn{6}{c|}{Accuracy = 0.814} &
  \multicolumn{6}{c|}{Accuracy = 0.646} &
  \multicolumn{6}{c|}{Accuracy = 0.503} \\ \cline{2-19} 
\multirow{-2}{*}{\textbf{BERT}} &
  \multicolumn{1}{l|}{0.814} &
  1.00 &
  0.898 &
  0.00 &
  {\color[HTML]{000000} 0.00} &
  0.00 &
  \multicolumn{1}{l|}{0.929} &
  0.676 &
  0.779 &
  0.018 &
  0.135 &
  0.036 &
  \multicolumn{1}{l|}{0.962} &
  0.511 &
  0.656 &
  0.009 &
  0.144 &
  0.009 \\ \hline
 &
  \multicolumn{6}{c|}{Accuracy = 0.559} &
  \multicolumn{6}{c|}{Accuracy = 0.973} &
  \multicolumn{6}{c|}{Accuracy = 0.985} \\ \cline{2-19} 

\multirow{-2}{*}{\color[HTML]{FE0000} {\ul \textbf{ELMo}}}&

  \multicolumn{1}{l|}{0.777} &
  0.642 &
  0.703 &
  0.11 &
  0.19 &
  0.14 &
  \multicolumn{1}{l|}{0.979} &
  0.993 &
  0.986 &
  {\color[HTML]{FE0000} {\ul \textbf{0.83}}} &
  0.64 &
  {\color[HTML]{FE0000} {\ul \textbf{0.72}}} &
  \multicolumn{1}{l|}{0.986} &
  0.99 &
  0.992 &
  0.88 &
  0.37 &
  0.52 \\ \hline
 &
  \multicolumn{6}{c|}{Accuracy = 0.851} &
  \multicolumn{6}{c|}{Accuracy = 0.946} &
  \multicolumn{6}{c|}{Accuracy = 0.98} \\ \cline{2-19} 
\multirow{-2}{*}{\textbf{Word2Vec}} &
  \multicolumn{1}{l|}{0.846} &
  0.999 &
  0.916 &
  0.98 &
  0.20 &
  0.33 &
  \multicolumn{1}{l|}{0.948} &
  0.998 &
  0.972 &
  0.60 &
  0.06 &
  0.12 &
  \multicolumn{1}{l|}{0.98} &
  1.00 &
  0.99 &
  1.00 &
  0.11 &
  0.19 \\ \hline
 &
  \multicolumn{6}{c|}{Accuracy = 0.599} &
  \multicolumn{6}{c|}{Accuracy = 0.953} &
  \multicolumn{6}{c|}{Accuracy = 0.984} \\ \cline{2-19} 
\multirow{-2}{*}{\textbf{GloVe}} &
  \multicolumn{1}{l|}{0.833} &
  0.635 &
  0.721 &
  0.22 &
  0.44 &
  0.29 &
  \multicolumn{1}{l|}{0.957} &
  0.995 &
  0.975 &
  0.73 &
  0.23 &
  0.35 &
  \multicolumn{1}{l|}{0.985} &
  0.999 &
  0.992 &
  {\color[HTML]{000000} 0.86} &
  {\color[HTML]{000000} 0.32} &
  0.46 \\ \hline
\multicolumn{19}{|c|}{\textbf{DSMs: Graph-based}} \\ \hline
 &
  \multicolumn{6}{c|}{Accuracy = 0.814} &
  \multicolumn{6}{c|}{Accuracy = 0.635} &
  \multicolumn{6}{c|}{Accuracy = 0.978} \\ \cline{2-19} 
\multirow{-2}{*}{\textbf{Text GCN}} &
  \multicolumn{1}{l|}{0.00} &
  0.00 &
  0.00 &
  0.81 &
  1.00 &
  0.90 &
  \multicolumn{1}{l|}{0.97} &
  0.633 &
  0.766 &
  0.095 &
  0.66 &
  0.165 &
  \multicolumn{1}{l|}{0.978} &
  1.00 &
  0.989 &
  0.00 &
  0.00 &
  0.00 \\ \hline
 &
  \multicolumn{6}{c|}{Accuracy = 0.677} &
  \multicolumn{6}{c|}{Accuracy = 0.64} &
  \multicolumn{6}{c|}{Accuracy = 0.458} \\ \cline{2-19} 
\multirow{-2}{*}{\textbf{VGCN-BERT}} &
  \multicolumn{1}{l|}{0.971} &
  0.622 &
  0.758 &
  {\color[HTML]{000000} 0.356} &
  0.917 &
  0.513 &
  \multicolumn{1}{l|}{0.985} &
  0.628 &
  0.767 &
  {\color[HTML]{000000} 0.117} &
  0.839 &
  0.205 &
  \multicolumn{1}{l|}{0.989} &
  0.451 &
  0.619 &
  0.031 &
  0.778 &
  0.06 \\ \hline
\multicolumn{19}{|c|}{\textbf{DSMs: Non-Graph + CNN}} \\ \hline
 &
  \multicolumn{6}{c|}{Accuracy = 0.96} &
  \multicolumn{6}{c|}{Accuracy = 0.643} &
  \multicolumn{6}{c|}{Accuracy = 0.99} \\ \cline{2-19} 

\multirow{-2}{*}{\color[HTML]{FE0000} {\ul \textbf{Word2Vec}}}&

  \multicolumn{1}{l|}{0.957} &
  0.984 &
  0.975 &
  {\color[HTML]{FE0000} {\ul \textbf{0.92}}} &
  {\color[HTML]{FE0000} {\ul \textbf{0.85}}} &
  {\color[HTML]{FE0000} {\ul \textbf{0.89}}} &
  \multicolumn{1}{l|}{0.977} &
  0.638 &
  0.772 &
  0.11 &
  {\color[HTML]{FE0000} {\ul \textbf{0.74}}} &
  0.19 &
  \multicolumn{1}{l|}{0.991} &
  0.99 &
  0.995 &
  {\color[HTML]{FE0000} {\ul \textbf{0.92}}} &
  {\color[HTML]{FE0000} {\ul \textbf{0.58}}} &
  {\color[HTML]{FE0000} {\ul \textbf{0.71}}} \\ \hline
 &
  \multicolumn{6}{c|}{Accuracy = 0.554} &
  \multicolumn{6}{c|}{Accuracy = 0.623} &
  \multicolumn{6}{c|}{Accuracy = 0.452} \\ \cline{2-19} 
\multirow{-2}{*}{\textbf{GloVe}} &
  \multicolumn{1}{l|}{0.775} &
  0.637 &
  0.699 &
  0.10 &
  0.19 &
  0.13 &
  \multicolumn{1}{l|}{0.941} &
  0.641 &
  0.763 &
  0.05 &
  0.32 &
  0.09 &
  \multicolumn{1}{l|}{0.962} &
  0.457 &
  0.62 &
  0.01 &
  0.21 &
  0.02 \\ \hline
\end{tabular}}
\vspace{-2.00mm}	
\caption{Study 1 \& 2: CoAID - News Articles (Temporal \& Text Length Dimension). Best results, as well as the optimal models, are highlighted in red.}
\label{CoAID-news}
\end{center}
\vspace{-2.00mm}	

\end{table*}


We evaluate the performance of the models based on the accuracy, precision, recall, and f1 score, with an emphasis on the misinformation class. For each experiment, we average the results for 10 runs. The experiments are done using Scikit-learn \cite{scikit-learn}, TensorFlow 2.0 \cite{tensorflow2015-whitepaper}, and PyTorch \cite{PyTorch} libraries. For creating the Word2Vec embeddings, we used the skip-gram model from the Gensim library \cite{gensim}. Finally, the GloVe embeddings are created using the model from \cite{glove}.

\paragraph{Results.} Table \ref{CoAID-Tweet} and Table \ref{CoAID-news} show the results from studies 1 and 2. 

\vspace{1.00mm}
From the results on the CoAID tweets, given in Table \ref{CoAID-Tweet}, we see that for the July tweet test dataset (Table \ref{CoAID-Tweet}), VGCN-BERT has the highest misinformation precision. However, misinformation recall and f1 scores for all models are poor. For September, the Text-GCN has outstanding performance for detecting misinformation, but its performance on true information is extremely poor. Other models perform badly on misinformation. For November, the GloVe-based transfer learning approach achieves excellent performance on both true information and misinformation, where precision, recall, and f1 scores are 1. Text-GCN also has decent scores on misinformation but fails to detect true information. The performance of BERT on both true information and misinformation is also good. However, we notice that no model performs well across three different test datasets. Thus, we see that \textbf{mixed-domain transfer learning is not robust when the context of the short-length data (tweets) changes}. This is also true for the DSM-based approach.

\begin{table*}[htb!]

\begin{center}
\scalebox{0.66}{
\begin{tabular}{|l|c|l|l|l|l|l|c|l|l|l|l|l|}
\hline
\multicolumn{1}{|c|}{\textbf{\begin{tabular}[c]{@{}c@{}}Train: May + July + September\\ Test: November\end{tabular}}} &
  \multicolumn{6}{c|}{\textbf{Tweets}} &
  \multicolumn{6}{c|}{\textbf{News Articles}} \\ \hline
\textbf{Model} &
  \multicolumn{3}{c|}{\textbf{True Information}} &
  \multicolumn{3}{c|}{\textbf{Misinformation}} &
  \multicolumn{3}{c|}{\textbf{True Information}} &
  \multicolumn{3}{c|}{\textbf{Misinformation}} \\ \hline
 &
  \multicolumn{1}{l|}{P} &
  R &
  F1 &
  P &
  R &
  F1 &
  \multicolumn{1}{l|}{P} &
  R &
  F1 &
  P &
  R &
  F1 \\ \hline
 &
  \multicolumn{6}{c|}{Accuracy = 0.928} &
  \multicolumn{6}{c|}{Accuracy = 0.992} \\ \cline{2-13} 
\multirow{-2}{*}{\textbf{Mixed-Domain Transfer Learning: BERT}} &
  \multicolumn{1}{l|}{1.00} &
  0.927 &
  0.962 &
  0.12 &
  {\color[HTML]{FE0000} {\ul \textbf{1.00}}} &
  0.22 &
  \multicolumn{1}{l|}{0.994} &
  0.998 &
  0.996 &
  {\color[HTML]{FE0000} {\ul \textbf{0.883}}} &
  {\color[HTML]{FE0000} {\ul \textbf{0.719}}} &
  {\color[HTML]{FE0000} {\ul \textbf{0.787}}} \\ \hline
 &
  \multicolumn{6}{c|}{Accuracy = 0.985} &
  \multicolumn{6}{c|}{Accuracy = 0.986} \\ \cline{2-13} 
\multirow{-2}{*}{\textbf{DSM (Non-Graph): Word2Vec + CNN}} &
  \multicolumn{1}{l|}{0.99} &
  0.995 &
  0.992 &
  {\color[HTML]{FE0000} {\ul \textbf{0.71}}} &
  {\color[HTML]{000000} 0.63} &
  {\color[HTML]{FE0000} {\ul \textbf{0.67}}} &
  \multicolumn{1}{l|}{0.992} &
  0.994 &
  0.993 &
  {\color[HTML]{000000} 0.71} &
  0.63 &
  {\color[HTML]{000000} 0.67} \\ \hline
\end{tabular}}
\vspace{-2.00mm}	
\caption{Study 3: CoAID Large Dataset (Dataset Size Dimension). Best results are highlighted in red.}
\label{CoAID-Big}
\end{center}
\vspace{-4.00mm}	

\end{table*}

Table \ref{CoAID-news} shows the results of CoAID news articles (lengthy text). For the July test dataset, both Text-GCN and  Word2Vec (DSM-based) achieve decent precision, recall, and f1 scores on misinformation. However, Text-GCN has extremely poor performance on true information. On the September data, ELMo exhibits the best misinformation precision, and f1 score, while Word2Vec (DSM-based) gives the best misinformation recall score. Both ELMo and Word2Vec perform well on the true information class as well. As for the November data, both transfer learning and DSM-based Word2Vec obtain optimal misinformation precision score and Word2Vec (DSM-based) obtains the highest f1 score. Besides, VGCN-BERT achieves the highest misinformation recall score. We notice that the DSM-based Word2Vec exhibits comparatively better performance across all test datasets. Thus, the \textbf{non-graph DSM+CNN can capture both global and local relationships from lengthy text relatively well}. The performance of the graph-based DSM approach on lengthy text is not as good as on short text. Also, BERT shows unreliable performance as it fails on the misinformation class.

Table \ref{CoAID-Big} shows the results of study 3, i.e., large-dataset-based experiments. The performance of DSM-based Word2Vec is consistent with its performance on the CoAID news articles data (Table \ref{CoAID-news}).  Its F1 score on tweets misinformation increases significantly compared to the small-data case (Table \ref{CoAID-Tweet}). \textbf{Thus, the non-graph DSM+CNN can capture both global and local relationships if we increase the size of short-length training data (i.e., tweets)}.

\begin{table*}[htb!]

\begin{center}
\scalebox{0.62}{
\begin{tabular}{|l|c|l|l|l|l|l|c|l|l|l|l|l|c|l|l|l|l|l|}
\hline
\multicolumn{1}{|c|}{\textbf{Train: May}} &
  \multicolumn{6}{c|}{\textbf{Test: July}} &
  \multicolumn{6}{c|}{\textbf{Test: September}} &
  \multicolumn{6}{c|}{\textbf{Test: November}} \\ \hline
\textbf{Model} &
  \multicolumn{3}{c|}{\textbf{True Information}} &
  \multicolumn{3}{c|}{\textbf{Misinformation}} &
  \multicolumn{3}{c|}{\textbf{True Information}} &
  \multicolumn{3}{c|}{\textbf{Misinformation}} &
  \multicolumn{3}{c|}{\textbf{True Information}} &
  \multicolumn{3}{c|}{\textbf{Misinformation}} \\ \hline
 &
  \multicolumn{1}{l|}{P} &
  R &
  F1 &
  P &
  R &
  F1 &
  \multicolumn{1}{l|}{P} &
  R &
  F1 &
  P &
  R &
  F1 &
  \multicolumn{1}{l|}{P} &
  R &
  F1 &
  P &
  R &
  F1 \\ \hline
\multicolumn{19}{|c|}{\textbf{Mixed-Domain Transfer Learning}} \\ \hline
 &
  \multicolumn{6}{c|}{Accuracy = 0.611} &
  \multicolumn{6}{c|}{Accuracy = 0.943} &
  \multicolumn{6}{c|}{Accuracy = 0.993} \\ \cline{2-19} 
\multirow{-2}{*}{\textbf{ALBERT}} &
  \multicolumn{1}{l|}{0.858} &
  0.625 &
  0.723 &
  0.25 &
  {\color[HTML]{000000} 0.547} &
  0.343 &
  \multicolumn{1}{l|}{0.96} &
  0.981 &
  0.97 &
  0.483 &
  0.298 &
  0.368 &
  \multicolumn{1}{l|}{0.996} &
  0.996 &
  0.996 &
  {\color[HTML]{FE0000} {\ul \textbf{0.842}}} &
  {\color[HTML]{FE0000} {\ul \textbf{0.842}}} &
  {\color[HTML]{FE0000} {\ul \textbf{0.842}}} \\ \hline
 &
  \multicolumn{6}{c|}{Accuracy = 0.936} &
  \multicolumn{6}{c|}{Accuracy = 0.630} &
  \multicolumn{6}{c|}{Accuracy = 0.980} \\ \cline{2-19} 
\multirow{-2}{*}{\textbf{RoBERTa}} &
  \multicolumn{1}{l|}{0.937} &
  0.987 &
  0.962 &
  {\color[HTML]{FE0000} {\ul \textbf{0.925}}} &
  {\color[HTML]{FE0000} {\ul \textbf{0.711}}} &
  {\color[HTML]{FE0000} {\ul \textbf{0.804}}} &
  \multicolumn{1}{l|}{0.952} &
  0.641 &
  0.766 &
  {\color[HTML]{000000} 0.068} &
  0.447 &
  {\color[HTML]{000000} 0.118} &
  \multicolumn{1}{l|}{0.991} &
  0.989 &
  0.990 &
  0.55 &
  0.579 &
  0.564 \\ \hline
 &
  \multicolumn{6}{c|}{Accuracy = 0.814} &
  \multicolumn{6}{c|}{Accuracy = 0.97} &
  \multicolumn{6}{c|}{Accuracy = 0.456} \\ \cline{2-19} 
\multirow{-2}{*}{\textbf{XLNET}} &
  \multicolumn{1}{l|}{0.81} &
  1.00 &
  0.90 &
  0.00 &
  0.00 &
  0.00 &
  \multicolumn{1}{l|}{0.97} &
  1.00 &
  0.98 &
  {\color[HTML]{FE0000} {\ul \textbf{0.89}}} &
  {\color[HTML]{FE0000} {\ul \textbf{0.53}}} &
  {\color[HTML]{FE0000} {\ul \textbf{0.67}}} &
  \multicolumn{1}{l|}{0.97} &
  0.46 &
  0.62 &
  0.01 &
  0.26 &
  0.02 \\ \hline
\end{tabular}}
\vspace{-2.00mm}	
\caption{Study 4: CoAID Large Dataset (Various-based Transformer Models). Best results are highlighted in red. \textbf{No single model performs well across three datasets}.}
\label{CoAID-bert-variants}
\vspace{-2.00mm}	
\end{center}

\end{table*}

\begin{figure*}[!ht]
  \centering
  \includegraphics[width=4.3in]{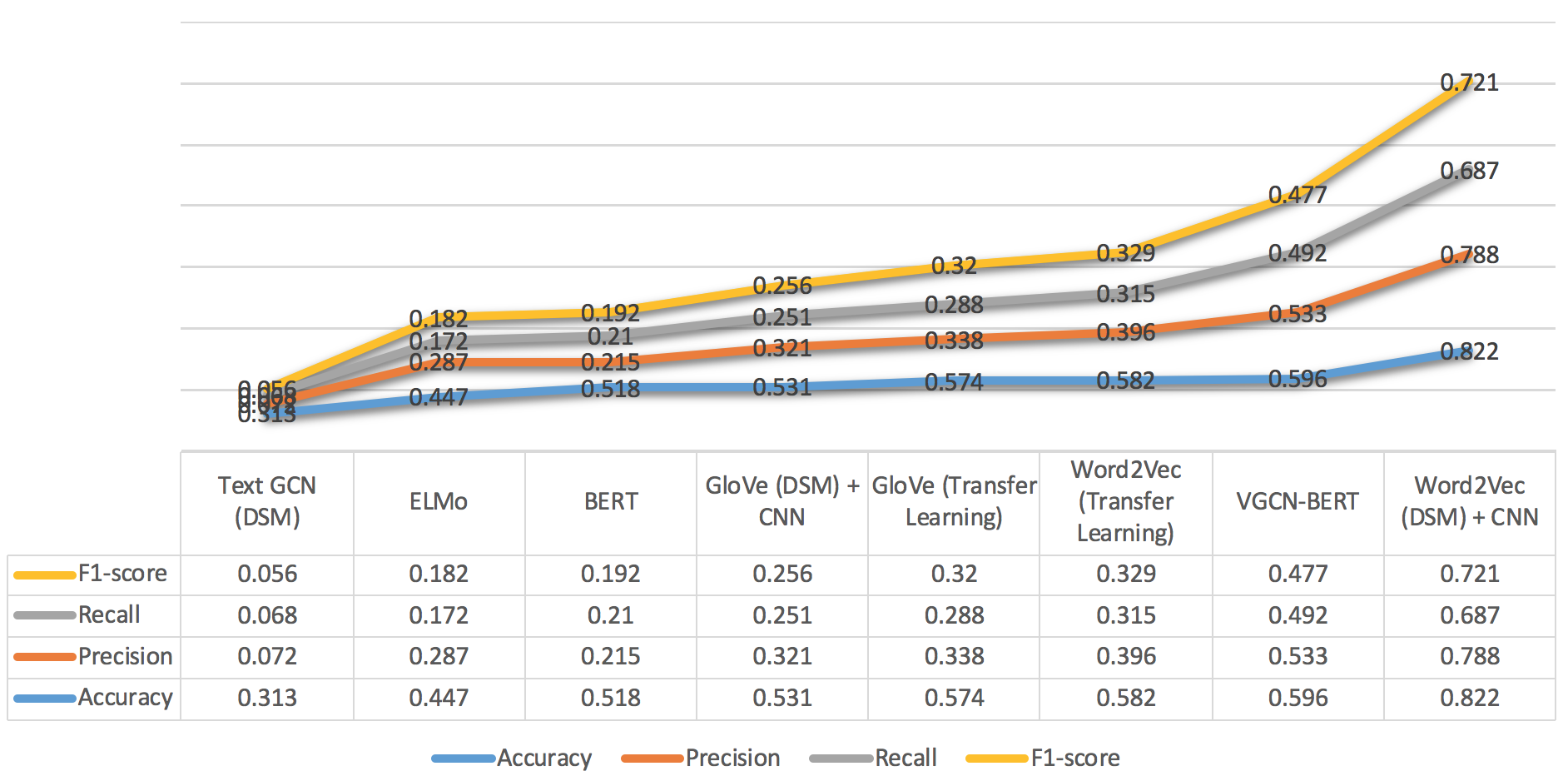}
  \vspace{-2.00mm}	
  \caption{Study 5: CMU-MisCov19 (Fine-grained classification).}
   \vspace{-5.00mm}	
  \label{fig:cmu-graph}
\end{figure*}

Table \ref{CoAID-bert-variants} shows the results obtained from study 4. For the July test dataset misinformation detection, RoBERTa achieves the best performance, while XLNet shows the worst performance. However, for September misinformation, we observe the exact opposite scenario. As for November misinformation, ALBERT achieves the best performance, while XLNet's performance is the worst. \textbf{No single Transformer-based model performs well on the three datasets.} These results corroborate our previous observation on the mixed-domain transfer learning approach, i.e., it is not robust when the context of the data changes. 


Figure \ref{fig:cmu-graph} shows the results obtained from study 5. We see that the mixed-domain transfer learning approach performs poorly on the fine-grained dataset. The only model that achieves decent performance is the non-graph DSM Word2Vec with CNN.

\begin{figure*}[!htb]
  \centering
  \includegraphics[width=3.2in]{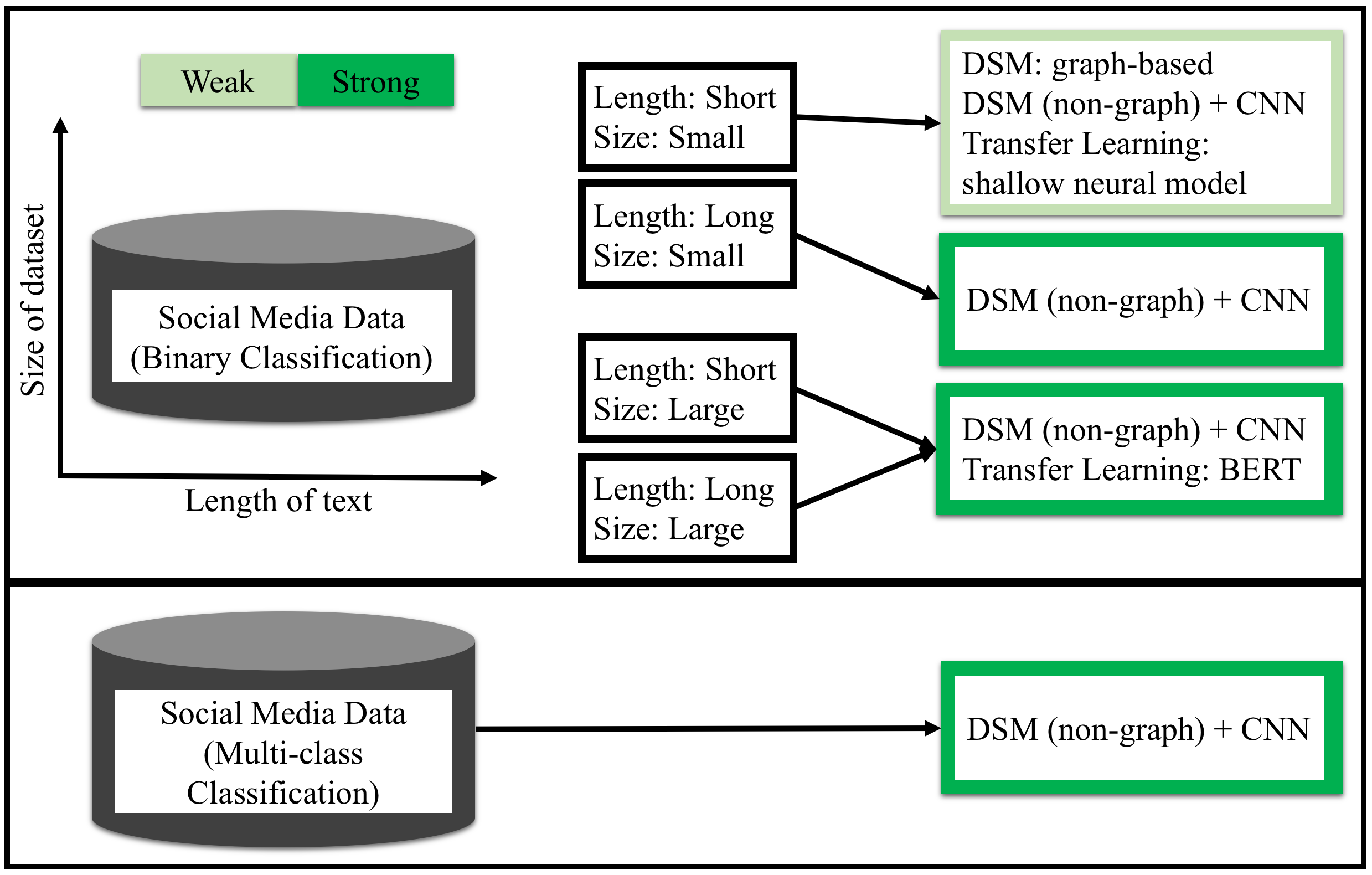}
  \vspace{-2.00 mm}
  \caption{A framework for COVID-19 online misinformation detection.}
  \label{fig:framework}
 \vspace{-4.00 mm}
\end{figure*}

\paragraph{Analysis.}

Based on the results obtained from the studies, we summarize our observations below. First, we discuss a model's generalizability for binary classification scenarios. Given the length of the text and the size of the dataset, we identify 4 cases.


\vspace{1.00mm}
\textbf{Case 1: Length=Short \& Size=Small} For case 1, we do not find a single best-performing model. For the tweet dataset, the following models perform slightly better: VGCN-BERT, GloVe (transfer learning and DSM-based), and Text GCN. There are two possible explanations for the poor performance of all models on the short-length tweet data. First, the number of test misinformation samples is significantly smaller. For example, in the 2020 July, September, and November tweet test datasets, the true information samples are larger than the misinformation samples by 46, 17, and 96 times, respectively. Second, the short length of the text and the small size of the training set might have influenced the scope of the context learning by the models. 



\vspace{1.00mm}
\textbf{Case 2: Length=Long \& Size=Small} For the news articles data, the best model is DSM Word2Vec+CNN for the July and November datasets. It achieved the highest precision and recall on the misinformation class. For the September dataset, the ELMo outperforms DSM Word2Vec+CNN.

\vspace{1.00mm}
\textbf{Case 3: Length=Short \& Size=Large} Both DSM Word2Vec+CNN and BERT-based transfer learning performed well. However, BERT's performance is not consistent. On the tweet dataset (short-length text), the precision of BERT is poor. It indicates that even with larger training data, BERT-based transfer learning does not provide an effective solution for short-length samples. One possible reason is that \textbf{although BERT is good at capturing the local relationships (e.g., word order), it does not do equally well on capturing the global relationships from short-length data}.

\vspace{1.00mm}
\textbf{Case 4: Length=Long \& Size=Large} Both DSM Word2Vec+CNN and BERT-based transfer learning perform well in this case. BERT's performance is slightly better. This indicates that Transformer-based models are suitable when target data is large and texts are lengthy.

\vspace{1.00mm}
The results from fine-grained classification show that DSM Word2Vec+CNN  outperforms other approaches by a large margin. Apart from the case of binary short length and small size dataset, DSM Word2Vec+CNN is shown to achieve the most effective solution.

One possible reason for the better generalization capability of the non-graph DSM+CNN-based approach is that its hierarchical feature-extraction mechanism is conducive for learning both the \textbf{local context} (the non-graph DSM, e.g., Word2Vec captures the local relationships of words in the target text) and the \textbf{global context} (the CNN learns global relationships from the word embeddings), which \textbf{validates our hypothesis}.

%

\vspace{2.00mm}
Based on the insights garnered from the above analysis, we draw the following conclusions, summarized in the framework in Figure \ref{fig:framework}.

\begin{itemize}

    \item The Transformer-based \textbf{mixed-domain transfer learning} approach is effective in limited cases. Also, its performance is not consistent.
    
    \item The \textbf{graph-based DSM} approach does not yield an effective solution in any of the cases. The VGCN-BERT that combines the benefits of Text GCN with BERT is not effective either.
    
    
     \item The \textbf{non-graph DSM + CNN} approach generalizes well across the last three cases.
    
\end{itemize}


Our study suffers from \textbf{some limitations}. The lack of labeled data narrowed the scope of our investigation. The data scarcity affected our study in two ways. First, due to the small size of the test data, we obtained noisy estimates for the short length and small size data. Second, we could not conduct a multi-dimensional study on the fine-grained classification problem.

%
%
%

\section{Conclusion}

\vspace{-1.50mm}
When an unanticipated pandemic like COVID-19 breaks out, various types of misinformation emerge and propagate at warp speed over online social media. For detecting such misinformation, NLP techniques require to capture the context of the discourse from its evolving narrative. We argue that irrespective of the success of the deep learning based mixed-domain transfer learning approach for solving various NLP tasks, it does not yield a generalizable solution. We emphasize the importance of learning the context (both local and global) directly from the target domain via the DSM-based approach. A feasible way to implement a DSM is to utilize shallow neural networks that capture the local relationships in the target data. Representations learned from this type of model can then be used by shallow CNNs to learn global relationships as high-level features. Thus, a combination of non-graph DSM and CNN may lend a more generalizable solution. We perform an extensive study using Twitter-based COVID-19 social media data that includes tweets and news articles referred to in the tweets. Our investigation is performed along the following dimensions of the data: temporal dimension (evolving context), length dimension (varying text length), size dimension (varying size of datasets), and classification-level dimension (binary vs. multi-class data). We show that the mixed-domain transfer learning approach does not always work well. We found the combination of the non-graph DSM (for capturing local relationships) and CNN (for extracting global relationships) to be a promising approach towards creating a generalizable solution for detecting COVID-19 online misinformation.


In the future, we plan to investigate the generalizability of the DSM models created using deep learning architectures such as BERT. 







\bibliography{references}

\begin{thebibliography}{54}
\expandafter\ifx\csname natexlab\endcsname\relax\def\natexlab#1{#1}\fi

\bibitem[{Abadi et~al.(2015)Abadi, Agarwal, Barham, Brevdo, Chen, Citro,
  Corrado, Davis, Dean, Devin, Ghemawat, Goodfellow, Harp, Irving, Isard, Jia,
  Jozefowicz, Kaiser, Kudlur, Levenberg, Man\'{e}, Monga, Moore, Murray, Olah,
  Schuster, Shlens, Steiner, Sutskever, Talwar, Tucker, Vanhoucke, Vasudevan,
  Vi\'{e}gas, Vinyals, Warden, Wattenberg, Wicke, Yu, and
  Zheng}]{tensorflow2015-whitepaper}
Mart\'{\i}n Abadi, Ashish Agarwal, Paul Barham, Eugene Brevdo, Zhifeng Chen,
  Craig Citro, Greg~S. Corrado, Andy Davis, Jeffrey Dean, Matthieu Devin,
  Sanjay Ghemawat, Ian Goodfellow, Andrew Harp, Geoffrey Irving, Michael Isard,
  Yangqing Jia, Rafal Jozefowicz, Lukasz Kaiser, Manjunath Kudlur, Josh
  Levenberg, Dandelion Man\'{e}, Rajat Monga, Sherry Moore, Derek Murray, Chris
  Olah, Mike Schuster, Jonathon Shlens, Benoit Steiner, Ilya Sutskever, Kunal
  Talwar, Paul Tucker, Vincent Vanhoucke, Vijay Vasudevan, Fernanda Vi\'{e}gas,
  Oriol Vinyals, Pete Warden, Martin Wattenberg, Martin Wicke, Yuan Yu, and
  Xiaoqiang Zheng. 2015.
\newblock \href {https://www.tensorflow.org/} {{TensorFlow}: Large-scale
  machine learning on heterogeneous systems}.
\newblock Software available from tensorflow.org.

\bibitem[{Bahdanau et~al.(2016)Bahdanau, Cho, and Bengio}]{Bahdanau:2016}
Dzmitry Bahdanau, Kyunghyun Cho, and Yoshua Bengio. 2016.
\newblock \href {http://arxiv.org/abs/1409.0473} {Neural machine translation by
  jointly learning to align and translate}.

\bibitem[{Beltagy et~al.(2019)Beltagy, Lo, and Cohan}]{beltagy2019scibert}
Iz~Beltagy, Kyle Lo, and Arman Cohan. 2019.
\newblock \href {https://doi.org/10.18653/v1/D19-1371} {{S}ci{BERT}: A
  pretrained language model for scientific text}.
\newblock In \emph{Proceedings of the 2019 Conference on Empirical Methods in
  Natural Language Processing and the 9th International Joint Conference on
  Natural Language Processing (EMNLP-IJCNLP)}, pages 3615--3620, Hong Kong,
  China. Association for Computational Linguistics.

\bibitem[{Brennen et~al.(2020)Brennen, Simon, Howard, and
  Nielsen}]{brennen_simon_howard_nielsen_2020}
Scott Brennen, Felix~N Simon, Philip~Kleis Howard, and Rasmus~undefined
  Nielsen. 2020.
\newblock \href
  {https://reutersinstitute.politics.ox.ac.uk/types-sources-and-claims-covid-19-misinformation}
  {Types, sources, and claims of covid-19 misinformation}.
\newblock \emph{Reuters Institute for the Study of Journalism}.

\bibitem[{Ciampaglia(2018)}]{Ciampagliapub.1093084646}
Giovanni~Luca Ciampaglia. 2018.
\newblock \href {https://doi.org/10.1007/s42001-017-0005-6} {Fighting fake
  news: a role for computational social science in the fight against digital
  misinformation}.
\newblock \emph{Journal of Computational Social Science}, 1(1):147--153.

\bibitem[{Ciampaglia et~al.(2018)Ciampaglia, Mantzarlis, Maus, and
  Menczer}]{Ciampaglia_Mantzarlis_Maus_Menczer_2018}
Giovanni~Luca Ciampaglia, Alexios Mantzarlis, Gregory Maus, and Filippo
  Menczer. 2018.
\newblock \href {https://doi.org/10.1609/aimag.v39i1.2783} {Research challenges
  of digital misinformation: Toward a trustworthy web}.
\newblock \emph{AI Magazine}, 39(1):65--74.

\bibitem[{Conneau et~al.(2017)Conneau, Schwenk, Barrault, and
  Lecun}]{Conneau:2017}
Alexis Conneau, Holger Schwenk, Lo{\"\i}c Barrault, and Yann Lecun. 2017.
\newblock \href {https://www.aclweb.org/anthology/E17-1104} {Very deep
  convolutional networks for text classification}.
\newblock In \emph{Proceedings of the 15th Conference of the {E}uropean Chapter
  of the Association for Computational Linguistics: Volume 1, Long Papers},
  pages 1107--1116, Valencia, Spain. Association for Computational Linguistics.

\bibitem[{Cui and Lee(2020)}]{cui2020coaid}
Limeng Cui and Dongwon Lee. 2020.
\newblock \href {http://arxiv.org/abs/2006.00885} {Coaid: {COVID-19} healthcare
  misinformation dataset}.
\newblock \emph{CoRR}, abs/2006.00885.

\bibitem[{Devlin et~al.(2019)Devlin, Chang, Lee, and
  Toutanova}]{devlin2019bert}
Jacob Devlin, Ming-Wei Chang, Kenton Lee, and Kristina Toutanova. 2019.
\newblock \href {https://doi.org/10.18653/v1/N19-1423} {{BERT}: Pre-training of
  deep bidirectional transformers for language understanding}.
\newblock In \emph{Proceedings of the 2019 Conference of the North {A}merican
  Chapter of the Association for Computational Linguistics: Human Language
  Technologies, Volume 1 (Long and Short Papers)}, pages 4171--4186,
  Minneapolis, Minnesota. Association for Computational Linguistics.

\bibitem[{{Glove-Python}(2016)}]{glove}
{Glove-Python}. 2016.
\newblock Glove-python.
\newblock \url{https://github.com/maciejkula/glove-python}.
\newblock Accessed: 2021-03-30.

\bibitem[{Goldberg and Hirst(2017)}]{Yoav:2017}
Yoav Goldberg and Graeme Hirst. 2017.
\newblock \emph{Neural Network Methods in Natural Language Processing}.
\newblock Morgan and Claypool Publishers.

\bibitem[{{Google Code}(2013)}]{google}
{Google Code}. 2013.
\newblock Google code archive - word2vec.
\newblock \url{https://code.google.com/archive/p/word2vec/}.
\newblock Accessed: 2021-04-05.

\bibitem[{Gu et~al.(2021)Gu, Tinn, Cheng, Lucas, Usuyama, Liu, Naumann, Gao,
  and Poon}]{Gu:2021}
Yu~Gu, Robert Tinn, Hao Cheng, Michael Lucas, Naoto Usuyama, Xiaodong Liu,
  Tristan Naumann, Jianfeng Gao, and Hoifung Poon. 2021.
\newblock \href {http://arxiv.org/abs/2007.15779} {Domain-specific language
  model pretraining for biomedical natural language processing}.

\bibitem[{Hamilton(2020)}]{hamilton_2020}
Isobel~Ashe Hamilton. 2020.
\newblock \href
  {https://www.businessinsider.com/77-phone-masts-fire-coronavirus-5g-conspiracy-theory-2020-5}
  {77 cell phone towers have been set on fire so far due to a weird coronavirus
  5g conspiracy theory}.
\newblock Business Insider.

\bibitem[{Hochreiter and Schmidhuber(1997)}]{Hochreiter:1997}
Sepp Hochreiter and J\"{u}rgen Schmidhuber. 1997.
\newblock \href {https://doi.org/10.1162/neco.1997.9.8.1735} {Long short-term
  memory}.
\newblock \emph{Neural Comput.}, 9(8):1735--1780.

\bibitem[{Hossain et~al.(2020)Hossain, Logan~IV, Ugarte, Matsubara, Young, and
  Singh}]{hossain-etal-2020-covidlies}
Tamanna Hossain, Robert~L. Logan~IV, Arjuna Ugarte, Yoshitomo Matsubara, Sean
  Young, and Sameer Singh. 2020.
\newblock \href {https://doi.org/10.18653/v1/2020.nlpcovid19-2.11}
  {{COVIDL}ies: Detecting {COVID}-19 misinformation on social media}.
\newblock In \emph{Proceedings of the 1st Workshop on {NLP} for {COVID}-19
  (Part 2) at {EMNLP} 2020}, Online. Association for Computational Linguistics.

\bibitem[{Hughes and Wojcik(2019)}]{hughes_wojcik_2020}
Adam Hughes and Stefan Wojcik. 2019.
\newblock \href
  {https://www.pewresearch.org/fact-tank/2019/08/02/10-facts-about-americans-and-twitter/}
  {10 facts about americans and twitter}.
\newblock Pew Research Center.

\bibitem[{Johnson et~al.(2020)Johnson, Velasquez, Jha, Niyazi, Leahy, Restrepo,
  Sear, Manrique, Lupu, Devkota, and Wuchty}]{johnson2020covid19}
N.~F. Johnson, N.~Velasquez, O.~K. Jha, H.~Niyazi, R.~Leahy, N.~Johnson
  Restrepo, R.~Sear, P.~Manrique, Y.~Lupu, P.~Devkota, and S.~Wuchty. 2020.
\newblock \href {http://arxiv.org/abs/2008.08513} {Covid-19 infodemic reveals
  new tipping point epidemiology and a revised $r$ formula}.

\bibitem[{Kim(2014)}]{Kim:2014}
Yoon Kim. 2014.
\newblock \href {https://doi.org/10.3115/v1/D14-1181} {Convolutional neural
  networks for sentence classification}.
\newblock In \emph{Proceedings of the 2014 Conference on Empirical Methods in
  Natural Language Processing ({EMNLP})}, pages 1746--1751, Doha, Qatar.
  Association for Computational Linguistics.

\bibitem[{Kipf and Welling(2017)}]{Kipf:2016}
Thomas~N. Kipf and Max Welling. 2017.
\newblock \href {https://openreview.net/forum?id=SJU4ayYgl} {{Semi-Supervised
  Classification with Graph Convolutional Networks}}.
\newblock In \emph{Proceedings of the 5th International Conference on Learning
  Representations}, ICLR 17.

\bibitem[{Lan et~al.(2020)Lan, Chen, Goodman, Gimpel, Sharma, and
  Soricut}]{lan2020albert}
Zhenzhong Lan, Mingda Chen, Sebastian Goodman, Kevin Gimpel, Piyush Sharma, and
  Radu Soricut. 2020.
\newblock \href {https://openreview.net/forum?id=H1eA7AEtvS} {Albert: A lite
  bert for self-supervised learning of language representations}.
\newblock In \emph{International Conference on Learning Representations}.

\bibitem[{Lazer et~al.(2018)Lazer, Baum, Benkler, Berinsky, Greenhill, Menczer,
  Metzger, Nyhan, Pennycook, Rothschild, Schudson, Sloman, Sunstein, Thorson,
  Watts, and Zittrain}]{Lazer1094}
David M.~J. Lazer, Matthew~A. Baum, Yochai Benkler, Adam~J. Berinsky, Kelly~M.
  Greenhill, Filippo Menczer, Miriam~J. Metzger, Brendan Nyhan, Gordon
  Pennycook, David Rothschild, Michael Schudson, Steven~A. Sloman, Cass~R.
  Sunstein, Emily~A. Thorson, Duncan~J. Watts, and Jonathan~L. Zittrain. 2018.
\newblock \href {https://doi.org/10.1126/science.aao2998} {The science of fake
  news}.
\newblock \emph{Science}, 359(6380):1094--1096.

\bibitem[{LeCun and Misra(2021)}]{LeCun:2021}
Yann LeCun and Ishan Misra. 2021.
\newblock \href
  {https://ai.facebook.com/blog/self-supervised-learning-the-dark-matter-of-intelligence/}
  {Self-supervised learning: The dark matter of intelligence}.
\newblock Facebook.

\bibitem[{Lee et~al.(2019)Lee, Yoon, Kim, Kim, Kim, So, and Kang}]{Lee:2019}
Jinhyuk Lee, Wonjin Yoon, Sungdong Kim, Donghyeon Kim, Sunkyu Kim, Chan~Ho So,
  and Jaewoo Kang. 2019.
\newblock \href {https://doi.org/10.1093/bioinformatics/btz682} {{BioBERT: a
  pre-trained biomedical language representation model for biomedical text
  mining}}.
\newblock \emph{Bioinformatics}, 36(4):1234--1240.

\bibitem[{Levy and Goldberg(2014)}]{Levy:2014}
Omer Levy and Yoav Goldberg. 2014.
\newblock Neural word embedding as implicit matrix factorization.
\newblock In \emph{NIPS}, pages 2177--2185.

\bibitem[{Liu et~al.(2016)Liu, Qiu, and Huang}]{Pengfei:2016}
Pengfei Liu, Xipeng Qiu, and Xuanjing Huang. 2016.
\newblock Recurrent neural network for text classification with multi-task
  learning.
\newblock In \emph{Proceedings of the Twenty-Fifth International Joint
  Conference on Artificial Intelligence}, IJCAI'16, page 2873?2879. AAAI Press.

\bibitem[{Liu et~al.(2019)Liu, Ott, Goyal, Du, Joshi, Chen, Levy, Lewis,
  Zettlemoyer, and Stoyanov}]{liu2019roberta}
Yinhan Liu, Myle Ott, Naman Goyal, Jingfei Du, Mandar Joshi, Danqi Chen, Omer
  Levy, Mike Lewis, Luke Zettlemoyer, and Veselin Stoyanov. 2019.
\newblock \href {http://arxiv.org/abs/1907.11692} {Roberta: A robustly
  optimized bert pretraining approach}.
\newblock Arxiv:1907.11692.

\bibitem[{Lu et~al.(2020)Lu, Du, and Nie}]{lu2020vgcnbert}
Zhibin Lu, Pan Du, and Jian-Yun Nie. 2020.
\newblock Vgcn-bert: Augmenting bert with graph embedding for text
  classification.
\newblock In \emph{Advances in Information Retrieval - 42nd European Conference
  on {IR} Research, {ECIR} 2020, Lisbon, Portugal, April 14-17, 2020,
  Proceedings, Part {I}}, volume 12035 of \emph{Lecture Notes in Computer
  Science}, pages 369--382. Springer.

\bibitem[{Maiya(2020)}]{maiya2020ktrain}
Arun~S. Maiya. 2020.
\newblock \href {http://arxiv.org/abs/2004.10703} {ktrain: A low-code library
  for augmented machine learning}.
\newblock \emph{arXiv preprint arXiv:2004.10703}.

\bibitem[{Memon and Carley(2020)}]{memon2020characterizing}
Shahan~Ali Memon and Kathleen~M. Carley. 2020.
\newblock \href {http://arxiv.org/abs/2008.00791} {Characterizing {COVID-19}
  misinformation communities using a novel twitter dataset}.
\newblock \emph{CoRR}, abs/2008.00791.

\bibitem[{Mikolov et~al.(2013)Mikolov, Chen, Corrado, and
  Dean}]{mikolov2013efficient}
Tomas Mikolov, Kai Chen, Greg Corrado, and Jeffrey Dean. 2013.
\newblock \href
  {http://dblp.uni-trier.de/db/journals/corr/corr1301.html#abs-1301-3781}
  {Efficient estimation of word representations in vector space}.
\newblock \emph{CoRR}, abs/1301.3781.

\bibitem[{Minaee et~al.(2021)Minaee, Kalchbrenner, Cambria, Nikzad, Chenaghlu,
  and Gao}]{Minaee:2021}
Shervin Minaee, Nal Kalchbrenner, Erik Cambria, Narjes Nikzad, Meysam
  Chenaghlu, and Jianfeng Gao. 2021.
\newblock \href {http://arxiv.org/abs/2004.03705} {Deep learning based text
  classification: A comprehensive review}.

\bibitem[{M{\"{u}}ller et~al.(2020)M{\"{u}}ller, Salath{\'{e}}, and
  Kummervold}]{Muller:2020}
Martin M{\"{u}}ller, Marcel Salath{\'{e}}, and Per~Egil Kummervold. 2020.
\newblock \href {http://arxiv.org/abs/2005.07503} {Covid-twitter-bert: {A}
  natural language processing model to analyse {COVID-19} content on twitter}.
\newblock \emph{CoRR}, abs/2005.07503.

\bibitem[{Neuman(2020)}]{neuman_2020}
Scott Neuman. 2020.
\newblock \href
  {https://www.npr.org/sections/coronavirus-live-updates/2020/03/24/820512107/man-dies-woman-hospitalized-after-taking-form-of-chloroquine-to-prevent-covid-19}
  {Man dies, woman hospitalized after taking form of chloroquine to prevent
  covid-19}.
\newblock \emph{NPR}.

\bibitem[{Pan and Yang(2009)}]{Pan:2009}
Sinno~Jialin Pan and Qiang Yang. 2009.
\newblock A survey on transfer learning.
\newblock \emph{IEEE Transactions on knowledge and data engineering},
  22(10):1345--1359.

\bibitem[{Paszke et~al.(2019)Paszke, Gross, Massa, Lerer, Bradbury, Chanan,
  Killeen, Lin, Gimelshein, Antiga, Desmaison, Kopf, Yang, DeVito, Raison,
  Tejani, Chilamkurthy, Steiner, Fang, Bai, and Chintala}]{PyTorch}
Adam Paszke, Sam Gross, Francisco Massa, Adam Lerer, James Bradbury, Gregory
  Chanan, Trevor Killeen, Zeming Lin, Natalia Gimelshein, Luca Antiga, Alban
  Desmaison, Andreas Kopf, Edward Yang, Zachary DeVito, Martin Raison, Alykhan
  Tejani, Sasank Chilamkurthy, Benoit Steiner, Lu~Fang, Junjie Bai, and Soumith
  Chintala. 2019.
\newblock \href
  {http://papers.neurips.cc/paper/9015-pytorch-an-imperative-style-high-performance-deep-learning-library.pdf}
  {Pytorch: An imperative style, high-performance deep learning library}.
\newblock In H.~Wallach, H.~Larochelle, A.~Beygelzimer, F.~d'Alch\'{e} Buc,
  E.~Fox, and R.~Garnett, editors, \emph{Advances in Neural Information
  Processing Systems 32}, pages 8024--8035. Curran Associates, Inc.

\bibitem[{Pedregosa et~al.(2011)Pedregosa, Varoquaux, Gramfort, Michel,
  Thirion, Grisel, Blondel, Prettenhofer, Weiss, Dubourg, Vanderplas, Passos,
  Cournapeau, Brucher, Perrot, and Duchesnay}]{scikit-learn}
F.~Pedregosa, G.~Varoquaux, A.~Gramfort, V.~Michel, B.~Thirion, O.~Grisel,
  M.~Blondel, P.~Prettenhofer, R.~Weiss, V.~Dubourg, J.~Vanderplas, A.~Passos,
  D.~Cournapeau, M.~Brucher, M.~Perrot, and E.~Duchesnay. 2011.
\newblock Scikit-learn: Machine learning in {P}ython.
\newblock \emph{Journal of Machine Learning Research}, 12:2825--2830.

\bibitem[{Pennington et~al.(2014)Pennington, Socher, and
  Manning}]{pennington-etal-2014-glove}
Jeffrey Pennington, Richard Socher, and Christopher Manning. 2014.
\newblock \href {https://doi.org/10.3115/v1/D14-1162} {{G}lo{V}e: Global
  vectors for word representation}.
\newblock In \emph{Proceedings of the 2014 Conference on Empirical Methods in
  Natural Language Processing ({EMNLP})}, pages 1532--1543, Doha, Qatar.
  Association for Computational Linguistics.

\bibitem[{Peters et~al.(2018)Peters, Neumann, Iyyer, Gardner, Clark, Lee, and
  Zettlemoyer}]{peters2018deep}
Matthew Peters, Mark Neumann, Mohit Iyyer, Matt Gardner, Christopher Clark,
  Kenton Lee, and Luke Zettlemoyer. 2018.
\newblock \href {https://doi.org/10.18653/v1/N18-1202} {Deep contextualized
  word representations}.
\newblock In \emph{Proceedings of the 2018 Conference of the North {A}merican
  Chapter of the Association for Computational Linguistics: Human Language
  Technologies, Volume 1 (Long Papers)}, pages 2227--2237, New Orleans,
  Louisiana. Association for Computational Linguistics.

\bibitem[{Peters et~al.(2019)Peters, Ruder, and Smith}]{peters2019tune}
Matthew~E. Peters, Sebastian Ruder, and Noah~A. Smith. 2019.
\newblock \href {https://doi.org/10.18653/v1/W19-4302} {To tune or not to tune?
  adapting pretrained representations to diverse tasks}.
\newblock In \emph{Proceedings of the 4th Workshop on Representation Learning
  for NLP (RepL4NLP-2019)}, pages 7--14, Florence, Italy. Association for
  Computational Linguistics.

\bibitem[{Qiu et~al.(2020)Qiu, Sun, Xu, Shao, Dai, and Huang}]{Qiu:2020}
Xipeng Qiu, Tianxiang Sun, Yige Xu, Yunfan Shao, Ning Dai, and Xuanjing Huang.
  2020.
\newblock \href {http://arxiv.org/abs/2003.08271} {Pre-trained models for
  natural language processing: A survey}.

\bibitem[{Radford and Sutskever(2018)}]{Radford:2018}
Alec Radford and Ilya Sutskever. 2018.
\newblock Improving language understanding by generative pre-training.
\newblock In \emph{arxiv}.

\bibitem[{{\v R}eh{\r u}{\v r}ek and Sojka(2010)}]{gensim}
Radim {\v R}eh{\r u}{\v r}ek and Petr Sojka. 2010.
\newblock {Software Framework for Topic Modelling with Large Corpora}.
\newblock In \emph{{Proceedings of the LREC 2010 Workshop on New Challenges for
  NLP Frameworks}}, pages 45--50, Valletta, Malta. ELRA.

\bibitem[{Schuster and Paliwal(1997)}]{Schuster:1997}
M.~Schuster and K.K. Paliwal. 1997.
\newblock \href {https://doi.org/10.1109/78.650093} {Bidirectional recurrent
  neural networks}.
\newblock \emph{Trans. Sig. Proc.}, 45(11):2673--2681.

\bibitem[{Srinivasan and Ribeiro(2019)}]{Srinivasan:2019}
Balasubramaniam Srinivasan and Bruno Ribeiro. 2019.
\newblock \href {http://arxiv.org/abs/1910.00452} {On the equivalence between
  node embeddings and structural graph representations}.
\newblock \emph{CoRR}, abs/1910.00452.

\bibitem[{Sutskever et~al.(2014)Sutskever, Vinyals, and Le}]{Sutskever:2014}
Ilya Sutskever, Oriol Vinyals, and Quoc~V Le. 2014.
\newblock \href
  {https://proceedings.neurips.cc/paper/2014/file/a14ac55a4f27472c5d894ec1c3c743d2-Paper.pdf}
  {Sequence to sequence learning with neural networks}.
\newblock In \emph{Advances in Neural Information Processing Systems},
  volume~27. Curran Associates, Inc.

\bibitem[{Tai et~al.(2015)Tai, Socher, and Manning}]{Tai:2015}
Kai~Sheng Tai, Richard Socher, and Christopher~D. Manning. 2015.
\newblock \href {http://arxiv.org/abs/1503.00075} {Improved semantic
  representations from tree-structured long short-term memory networks}.
\newblock Cite arxiv:1503.00075Comment: Accepted for publication at ACL 2015.

\bibitem[{Tendle and Hasan(2021)}]{Tendle:2021}
Atharva Tendle and Mohammad~Rashedul Hasan. 2021.
\newblock \href {https://doi.org/https://doi.org/10.1016/j.mlwa.2021.100124} {A
  study of the generalizability of self-supervised representations}.
\newblock \emph{Machine Learning with Applications}, 6:100124.

\bibitem[{Vaswani et~al.(2017)Vaswani, Shazeer, Parmar, Uszkoreit, Jones,
  Gomez, Kaiser, and Polosukhin}]{Vaswani:2017}
Ashish Vaswani, Noam Shazeer, Niki Parmar, Jakob Uszkoreit, Llion Jones,
  Aidan~N. Gomez, Lukasz Kaiser, and Illia Polosukhin. 2017.
\newblock \href
  {https://proceedings.neurips.cc/paper/2017/hash/3f5ee243547dee91fbd053c1c4a845aa-Abstract.html}
  {Attention is all you need}.
\newblock In \emph{Advances in Neural Information Processing Systems 30: Annual
  Conference on Neural Information Processing Systems 2017, December 4-9, 2017,
  Long Beach, CA, {USA}}, pages 5998--6008.

\bibitem[{Wang et~al.(2017)Wang, Wang, Zhang, and Yan}]{Wang:2017}
Jin Wang, Zhongyuan Wang, Dawei Zhang, and Jun Yan. 2017.
\newblock Combining knowledge with deep convolutional neural networks for short
  text classification.
\newblock In \emph{Proceedings of the 26th International Joint Conference on
  Artificial Intelligence}, IJCAI'17, pages 2915--2921. AAAI Press.

\bibitem[{Wolf et~al.(2020)Wolf, Debut, Sanh, Chaumond, Delangue, Moi, Cistac,
  Rault, Louf, Funtowicz, Davison, Shleifer, von Platen, Ma, Jernite, Plu, Xu,
  Scao, Gugger, Drame, Lhoest, and Rush}]{wolf-etal-2020-transformers}
Thomas Wolf, Lysandre Debut, Victor Sanh, Julien Chaumond, Clement Delangue,
  Anthony Moi, Pierric Cistac, Tim Rault, Rémi Louf, Morgan Funtowicz, Joe
  Davison, Sam Shleifer, Patrick von Platen, Clara Ma, Yacine Jernite, Julien
  Plu, Canwen Xu, Teven~Le Scao, Sylvain Gugger, Mariama Drame, Quentin Lhoest,
  and Alexander~M. Rush. 2020.
\newblock \href {https://www.aclweb.org/anthology/2020.emnlp-demos.6}
  {Transformers: State-of-the-art natural language processing}.
\newblock In \emph{Proceedings of the 2020 Conference on Empirical Methods in
  Natural Language Processing: System Demonstrations}, pages 38--45, Online.
  Association for Computational Linguistics.

\bibitem[{Yang et~al.(2019)Yang, Dai, Yang, Carbonell, Salakhutdinov, and
  Le}]{yang2020xlnet}
Zhilin Yang, Zihang Dai, Yiming Yang, Jaime Carbonell, Russ~R Salakhutdinov,
  and Quoc~V Le. 2019.
\newblock \href
  {https://proceedings.neurips.cc/paper/2019/file/dc6a7e655d7e5840e66733e9ee67cc69-Paper.pdf}
  {Xlnet: Generalized autoregressive pretraining for language understanding}.
\newblock In \emph{Advances in Neural Information Processing Systems},
  volume~32. Curran Associates, Inc.

\bibitem[{Yao et~al.(2019)Yao, Mao, and Luo}]{yao2018graph}
Liang Yao, Chengsheng Mao, and Yuan Luo. 2019.
\newblock \href {https://doi.org/10.1609/aaai.v33i01.33017370} {Graph
  convolutional networks for text classification}.
\newblock \emph{Proceedings of the AAAI Conference on Artificial Intelligence},
  33(01):7370--7377.

\bibitem[{Zhou et~al.(2016)Zhou, Qi, Zheng, Xu, Bao, and Xu}]{Zhou:2016}
Peng Zhou, Zhenyu Qi, Suncong Zheng, Jiaming Xu, Hongyun Bao, and Bo~Xu. 2016.
\newblock \href
  {http://dblp.uni-trier.de/db/conf/coling/coling2016.html#ZhouQZXBX16} {Text
  classification improved by integrating bidirectional lstm with
  two-dimensional max pooling.}
\newblock In \emph{COLING}, pages 3485--3495. ACL.

\end{thebibliography}
\bibliographystyle{acl_natbib}


\section*{Appendix}

In this section, first, we discuss the related work. Then, we present an analysis of the dataset. Finally, we report the experimental setting and training statistics.

\section{Related Work}


Solving Natural Language Processing (NLP) tasks using Deep Learning (DL) based models is a challenging venture. Unlike computer vision problems in which deep learning supervised model can learn expressive representations directly from raw pixels of the input data while performing a discrimination task, the deep learning based supervised NLP systems cannot use raw text input while solving NLP tasks. The text input data needs to be encoded with latent representations or embeddings. These embeddings are learned by neural models from general-purpose unlabeled data using the self-supervised learning approach \cite{Tendle:2021}. The embeddings must capture the multi-dimensional relationships of the text components, which are non-contextual and contextual relationships. The non-contextual relationship includes syntactic relationships and semantic relationships. On the other hand, the contextual relationship includes dynamic representations of words, which requires embeddings to capture the local and global relationships of the words. 


Sequence DL models such as Recurrent Neural Network (RNN) have been used to learn the local context of a word in sequential order \cite{Sutskever:2014}. RNNs process text as a sequence of words for capturing word dependencies and text structures. However, they suffer from two limitations. First, they are unable to create good representations due to the uni-directional processing \cite{peters2018deep}. Second, these models struggle with capturing long-term dependency \cite{Hochreiter:1997}. These two issues were partially resolved by introducing the bi-directional LSTM model \cite{Schuster:1997}. This model was combined with two-dimensional max-pooling in \cite{Zhou:2016} for capturing text features. In addition to this type of chain-structured LSTM, tree-structured LSTM such as the Tree-LSTM model was developed for learning rich semantic representations \cite{Tai:2015}. Irrespective of the progress harnessed by RNN-based models, they do not perform well in capturing global relationships (i.e., long-term dependencies) among the words of the source text. Also, training this type of model on large data is inefficient.

An efficient approach for some NLP tasks such as text classification is a shallow Convolutional Neural Network (CNN) with a one-dimensional convolutional kernel \cite{Kim:2014}. This model is good at capturing local patterns such as key phrases in the text. However, it does not work effectively if the weights of the input layer are initialized randomly \cite{Kim:2014}. It was shown to be effective only in transfer learning in which, first, word embeddings are created using a self-supervised pre-trained model (PTM) such as Word2Vec \cite{mikolov2013efficient}, then the CNN uses its single layer of convolution on top of the word embeddings to learn high-level representations.

The use of PTMs for mixed-domain transfer learning ushered in a new era in NLP \cite{Qiu:2020}. The PTMs are created from general-purpose unlabeled data by using the self-supervised learning technique. In general, the SSL technique learns representations by predicting a hidden property of the input from the observable properties \cite{LeCun:2021}. Two types of PTMs are used in NLP: (i) PTMs that are feature extractors, i.e., learn word embeddings \cite{mikolov2013efficient, pennington-etal-2014-glove, peters2018deep}, which are used as input to another model for solving a downstream NLP task \cite{Kim:2014}, and (ii) PTMs that learn language models and the same PTM is adapted (fine-tuned) for solving downstream NLP tasks \cite{devlin2019bert}. The feature extractor PTMs such as Word2Vec  \cite{mikolov2013efficient}, GloVe  \cite{pennington-etal-2014-glove}, and ELMo \cite{peters2018deep} are based on both shallow and deep neural network architectures. While shallow Word2Vec and GloVe models learn non-contextual word embeddings from unlabeled source data, the deep ELMo model  is good for creating contextual embeddings. Features learned from these PTMs are used as input to another neural network for solving a downstream NLP task using labeled target data.

Both Word2Vec and GloVe learn word embeddings from their co-occurrence information. While Word2Vec leverages co-occurrence within the local context, GloVe utilizes global word-to-word co-occurrence counts from the entire corpus. Word2Vec is a shallow feed-forward neural network-based predictive model that learns embeddings of the words while improving their predictions within the local context. On the other hand, GloVe is a count-based model that applies dimensionality reduction on the co-occurrence count matrix for learning word embeddings. These two models are good at capturing syntactic as well as semantic relationships. Although they can capture some global relationships between words in a text \cite{Levy:2014, Srinivasan:2019}, their embeddings are context-independent. Thus, these two models are not good at language modeling. A language model can predict the next word in the sequence given the words that precede it \cite{Yoav:2017}, which requires it to capture the context of the text. The deep architecture feature extractor PTM ELMo (Embeddings from Language Models) \cite{peters2018deep} learns contextualized word embeddings, i.e., it maps a word to different embedding vectors depending on their context. It uses two LSTMs in the forward and backward directions to encode the context of the words. The main limitation of this deep PTM is that it is computationally complex due to its sequential processing of text. Thus, it is prohibitively expensive to train using a very large text corpus. Another limitation of this model, which also applies to feature extractor PTMs in general, is that for solving downstream NLP tasks we need to train the entire model, except for the input embedding layer, from scratch.

The above two limitations of the feature extractor PTMs are addressed by a very deep architecture-based Transformer model \cite{Vaswani:2017}. Unlike the feature extractor sequential PTMs, Transformer is a non-sequential model that uses self-attention \cite{Bahdanau:2016} to compute an attention score for capturing the influence of every word on other words in a sentence or document. This process is parallelized, which enables training deep Transformer models efficiently using very large text corpus such as Wikipedia and BookCorpus \cite{devlin2019bert} as well as web crawls \cite{liu2019roberta}.

There are two main types of Transformer-based deep PTMs: autoregressive and autoencoding. The OpenAI GPT (Generative Pre-training) \cite{Radford:2018} is an autoregressive model that learns embeddings by predicting words based on previous predictions. Specifically, it is a uni-directional model that predicts words sequentially in a text. On the other hand, BERT \cite{devlin2019bert} utilizes the autoencoding technique based on bi-directional context modeling. Specifically, for training, it uses a masked language modeling (MLM) task. The MLM randomly masks some tokens in a text sequence, then it predicts the masked tokens by learning the encoding vectors. 

Variants of BERT such as RoBERTa  \cite{liu2019roberta} and ALBERT \cite{lan2020albert} were proposed to improve its effectiveness as well as efficiency. RoBERTa (Robustly optimized BERT) improves the effectiveness of BERT by using several strategies that include the following. It trains the model longer using more data, lengthy input, and larger batches. It uses a dynamic masking strategy and removes BERT's Next Sentence Prediction (NSP) task. ALBERT (A Lite BERT) improves the efficiency of BERT by employing fewer parameters, which increases its training speed. 

There have been attempts such as in XLNet \cite{yang2020xlnet} to integrate the strengths of the autoregressive and autoencoding Transformer techniques. XLNet is an autoregressive model that uses a permutation language modeling objective. This allows XLNet to retain the advantages of autoregressive models while leveraging the benefit of the autoencoding models, i.e., to capture the bi-directional context.

Irrespective of the state-of-the-art (SOTA) performance of the deep PTM based mixed-domain transfer learning approach on many NLP tasks \cite{Minaee:2021}, this approach is not suitable for detecting misinformation from COVID-19 social media data. It generalizes poorly when the domain of the source dataset used to create the PTMs is significantly different from that of the target dataset \cite{peters2019tune}. One solution to this generalizability problem is to create a PTM using data that shares context similar to the target domain, i.e., pre-train a domain-specific model (DSM). This type of model encodes the context of the target domain more effectively to provide a generalizable solution for the downstream task \cite{beltagy2019scibert, Lee:2019, Gu:2021}. However, pre-training a deep architecture-based DSM (e.g., BERT) for the COVID-19 misinformation detection task in a timely fashion could be infeasible as it requires collecting a large amount of COVID-19 social media data, which must cover the diverse landscape of COVID-19 misinformation. While there was an effort to create such a deep DSM using COVID-19 tweets in \cite{Muller:2020}, capturing the dynamic context of the pandemic requires the collection of various types of social media data at a large scale. 

Thus, to create DSMs for the COVID-19 domain using quickly collectible small data, shallow architecture based PTMs such as Word2Vec and GloVe are suitable. However, as mentioned earlier, these PTMs are context-independent and are not good at capturing the global relationships well. To compensate for these shortcomings, we used the extracted features from the Word2Vec and GloVe DSMs  for training a one-dimensional convolutional kernel-based CNN similar to the shallow architecture given in \cite{Kim:2014}. The CNN learns global relationships by extracting local patterns in a hierarchical fashion by convolving over the word embeddings. 

Another type of DSM we used is graph-based that leverages the linguistic-aware graph structure of the text for learning contextual representations, then uses those representations to solve a downstream NLP task. The main intuition driving the graph-based technique is that by modeling the vocabulary graph, it will be possible to encode global relationships in the embeddings. Text GCN (Text Graph Convolutional Network) \cite{yao2018graph} is a graph-based model that explicitly captures the global term-co-occurrence information by leveraging the graph structure of the text. It models the global word co-occurrence by incorporating edges between words as well as edges between a document and a word. Word-word edges are created by using word co-occurrence information and word-document edges are created by using word frequency and word-document frequency. Its input is a one-hot vector representation of every word in the document, which is used to create a heterogeneous text graph that has word nodes and document nodes. These are fed into a two-layer GCN (Graph Convolutional Network) \cite{Kipf:2016} that turns document classification into a node classification problem.

Although Text GCN is good at convolving the global information in the graph, it does not take into account local information such as word orders. To address this issue, Text GCN was combined with BERT, which is good at capturing local information. The resulting model is VGCN-BERT \cite{lu2020vgcnbert}. BERT captures the local context by focusing on local word sequences, but it is not good at capturing global information of a text. It learns representations from a sentence or a document. However, it does not take into account the knowledge of the vocabulary. Thus its language model may be incomplete. On the other hand, Text GCN captures the global vocabulary information. The VGCN-BERT aims to capture both local and global relationships by integrating GCN with BERT. Both the graph embeddings and word embeddings are fed into a self-attention encoder in BERT.  When the classifier is trained, these two types of embeddings interact with each other through the self-attention mechanism. As a consequence, the classifier creates representations by fusing global information with local information in a guided fashion.

\section{Dataset}

We describe the sample distribution of both the CoAID (binary) and CMU datasets. Then, we analyze context evolution in the CoAID new articles dataset.

\begin{figure}[!htb]
  \centering
\includegraphics[width=\linewidth]{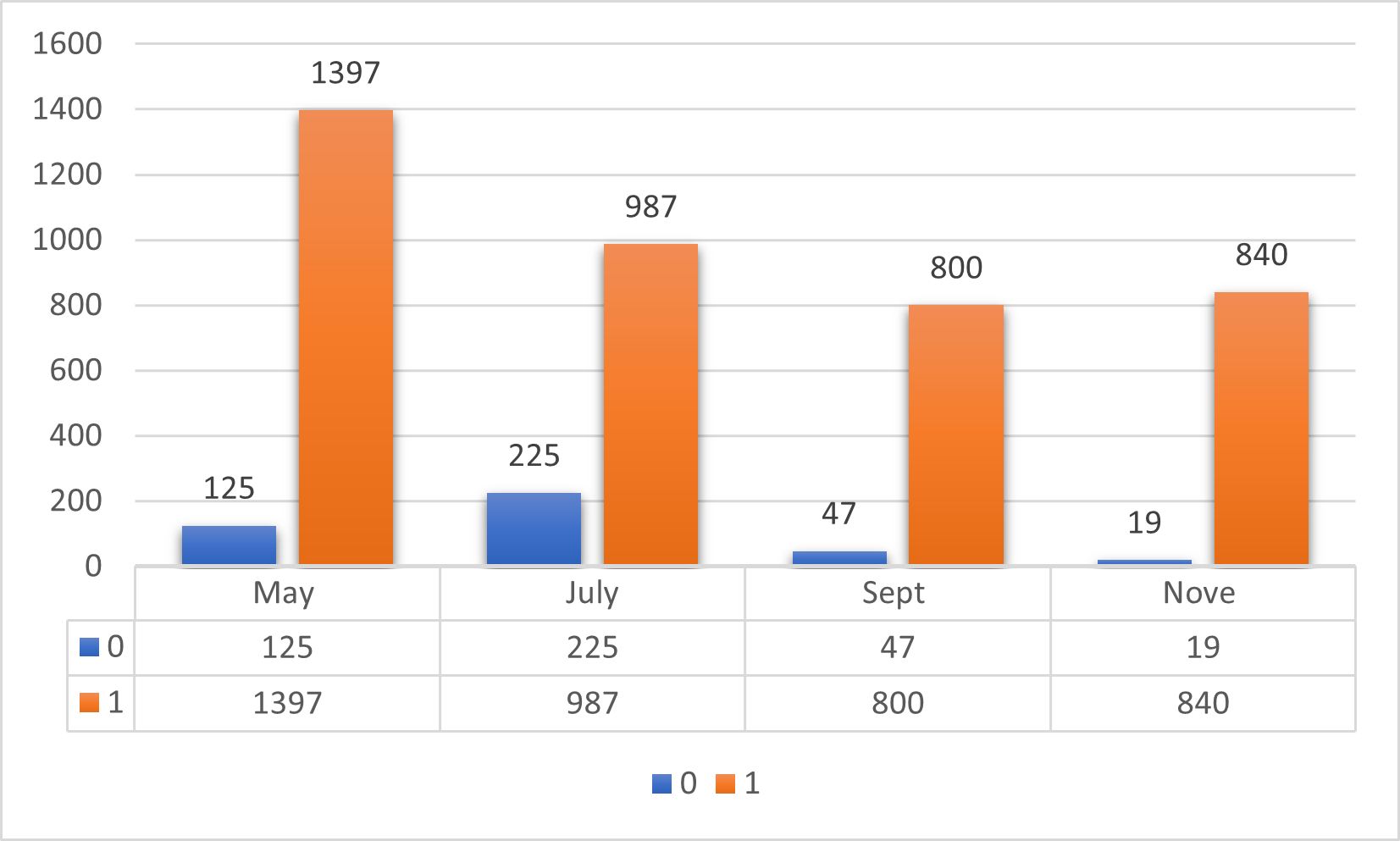}
  \caption{CoAID Tweets: Sample distribution (0: misinformation, 1: true information).}
  \label{fig:coaid-tweet}
\end{figure}


\label{coaid}
\begin{figure}[!htb]
  \centering
  \includegraphics[width=\linewidth]{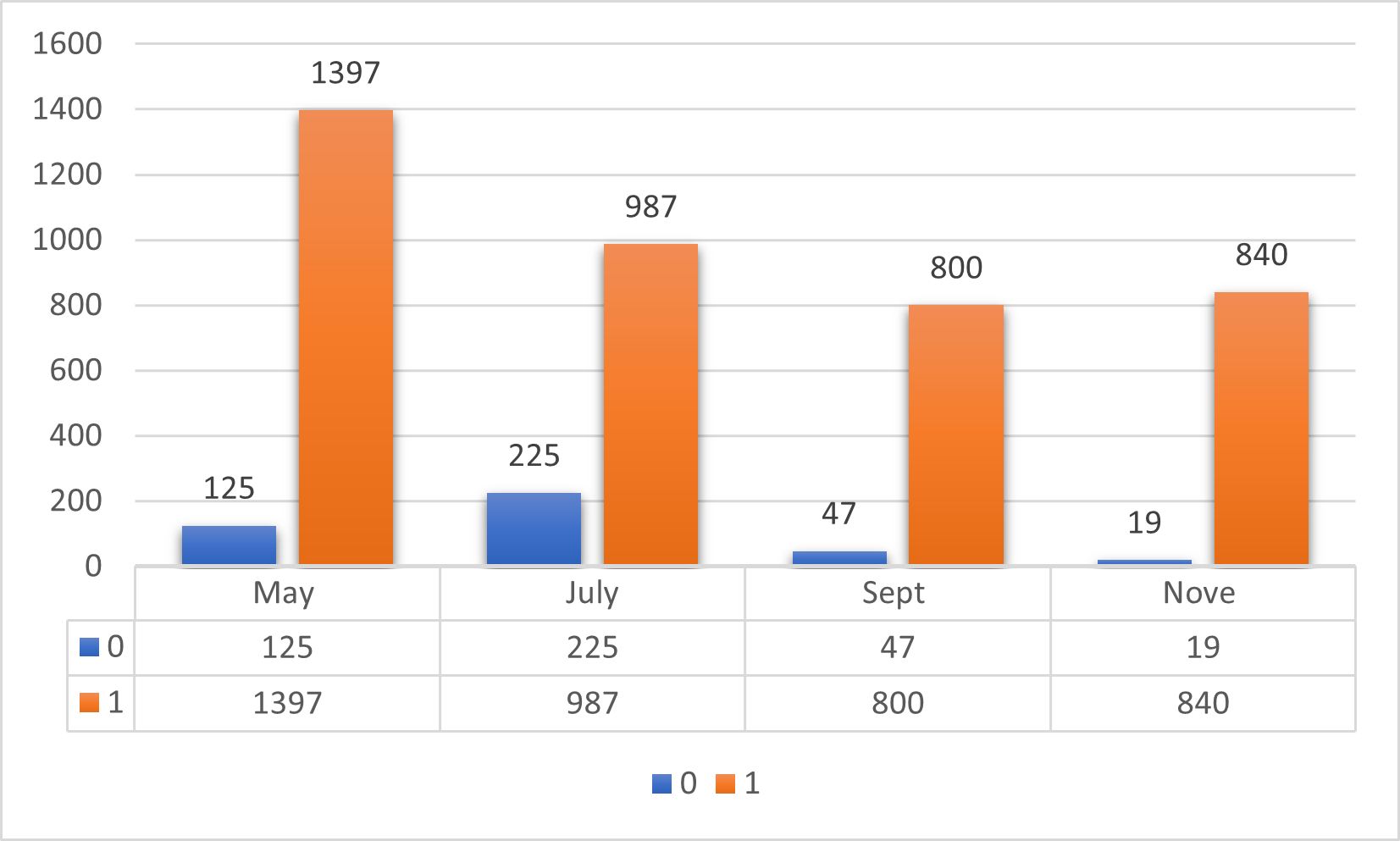}
  \caption{CoAID News Articles: Sample distribution (0: misinformation, 1: true information).}
  \label{fig:coaid-news}
\end{figure}

\begin{figure}[htb!]
  \centering
  \includegraphics[width=\linewidth]{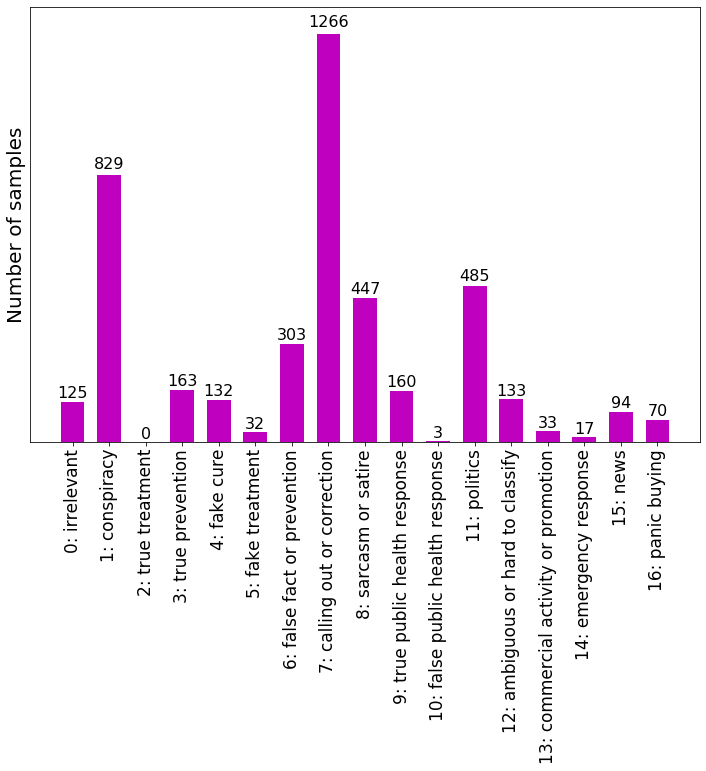} 
  \caption{CMU-MisCov19 Dataset: Sample distribution.}
  \label{fig:cmu}
\end{figure}

\begin{figure*}[htb!]
    \begin{center}
        \subfigure[May]{\label{news-may}\includegraphics[width=2.20in]{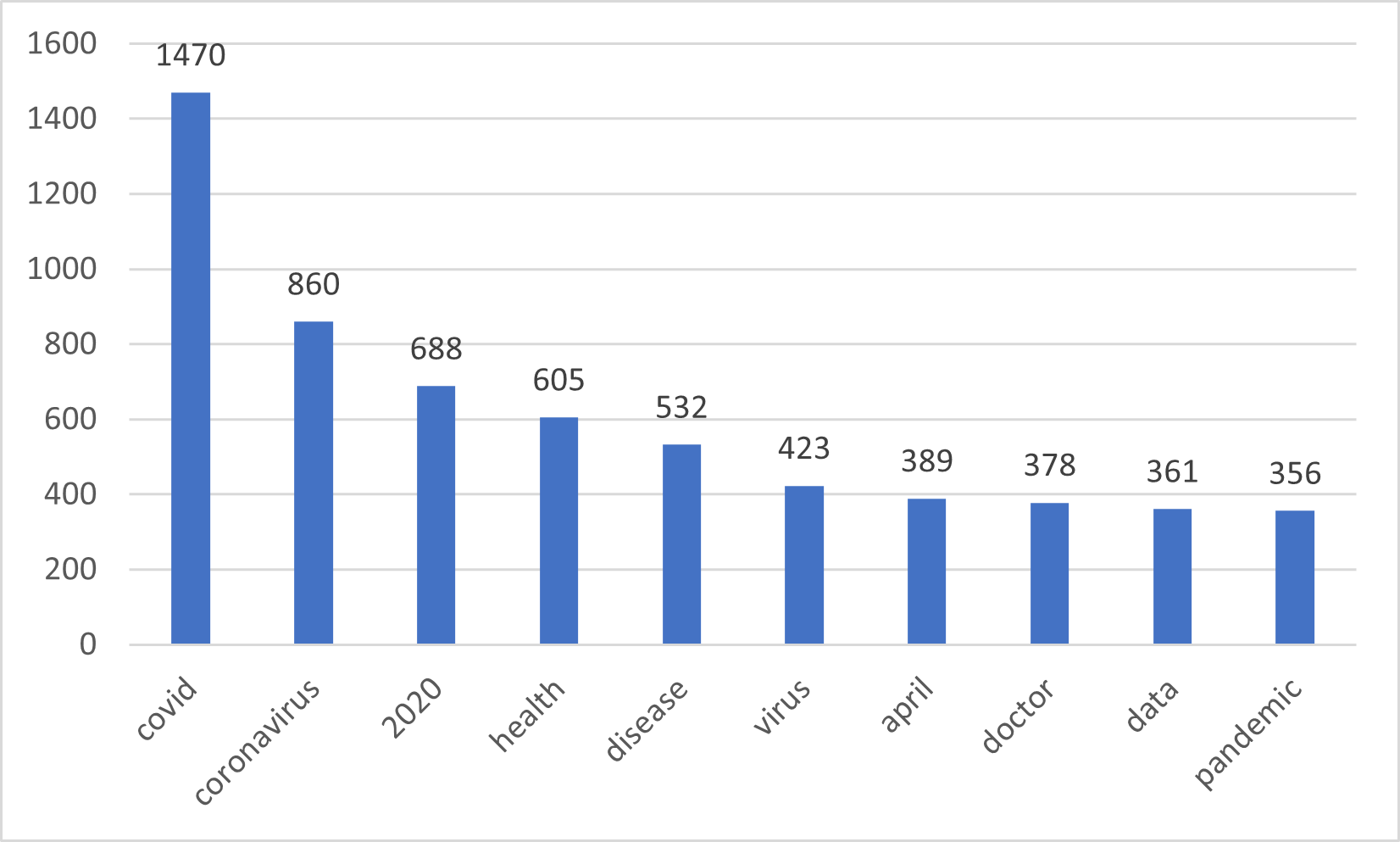}} \quad	
        \subfigure[July]{\label{news-july}\includegraphics[width=2.20in]{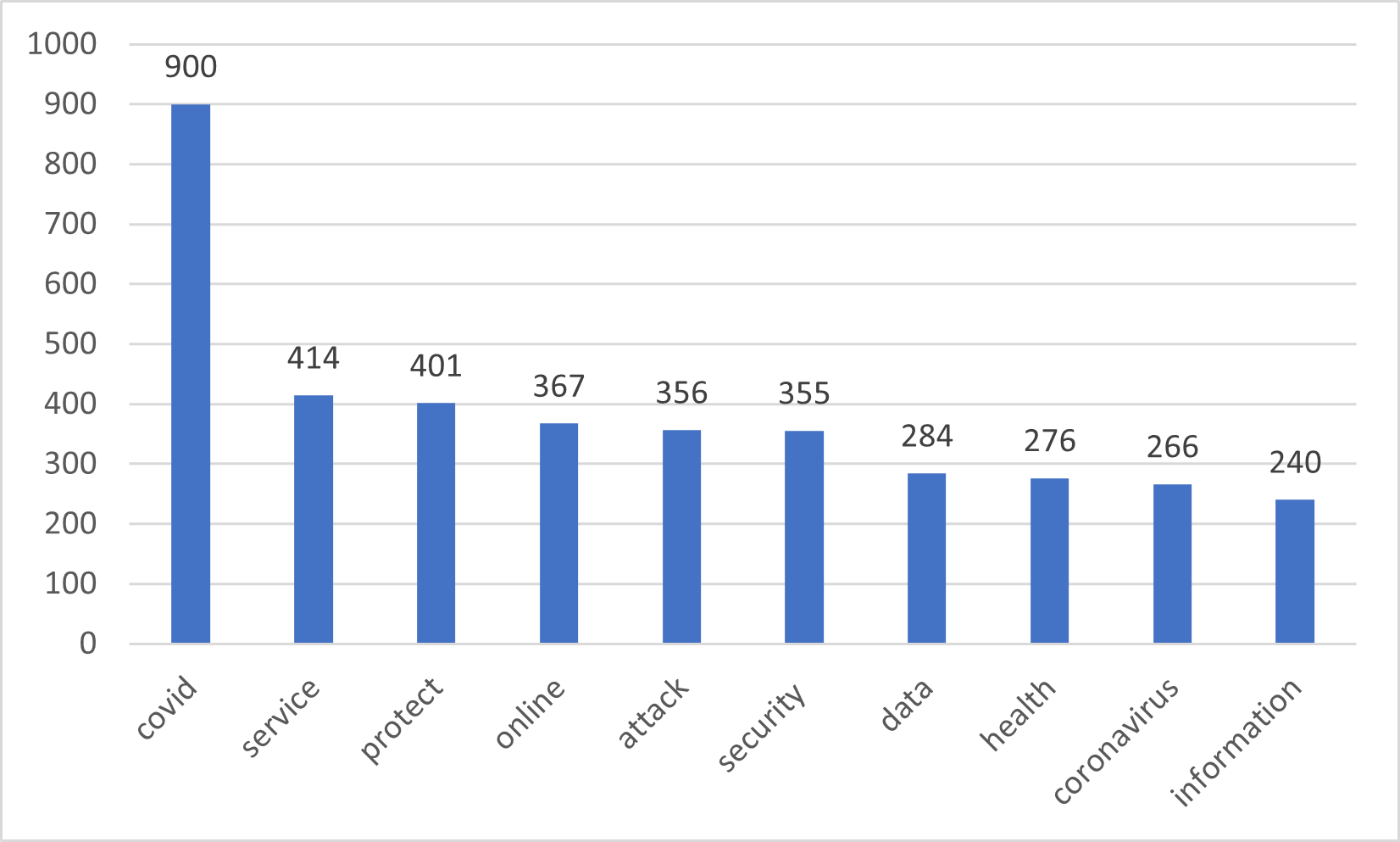}} \\	
        \subfigure[September]{\label{news-sept}\includegraphics[width=2.20in]{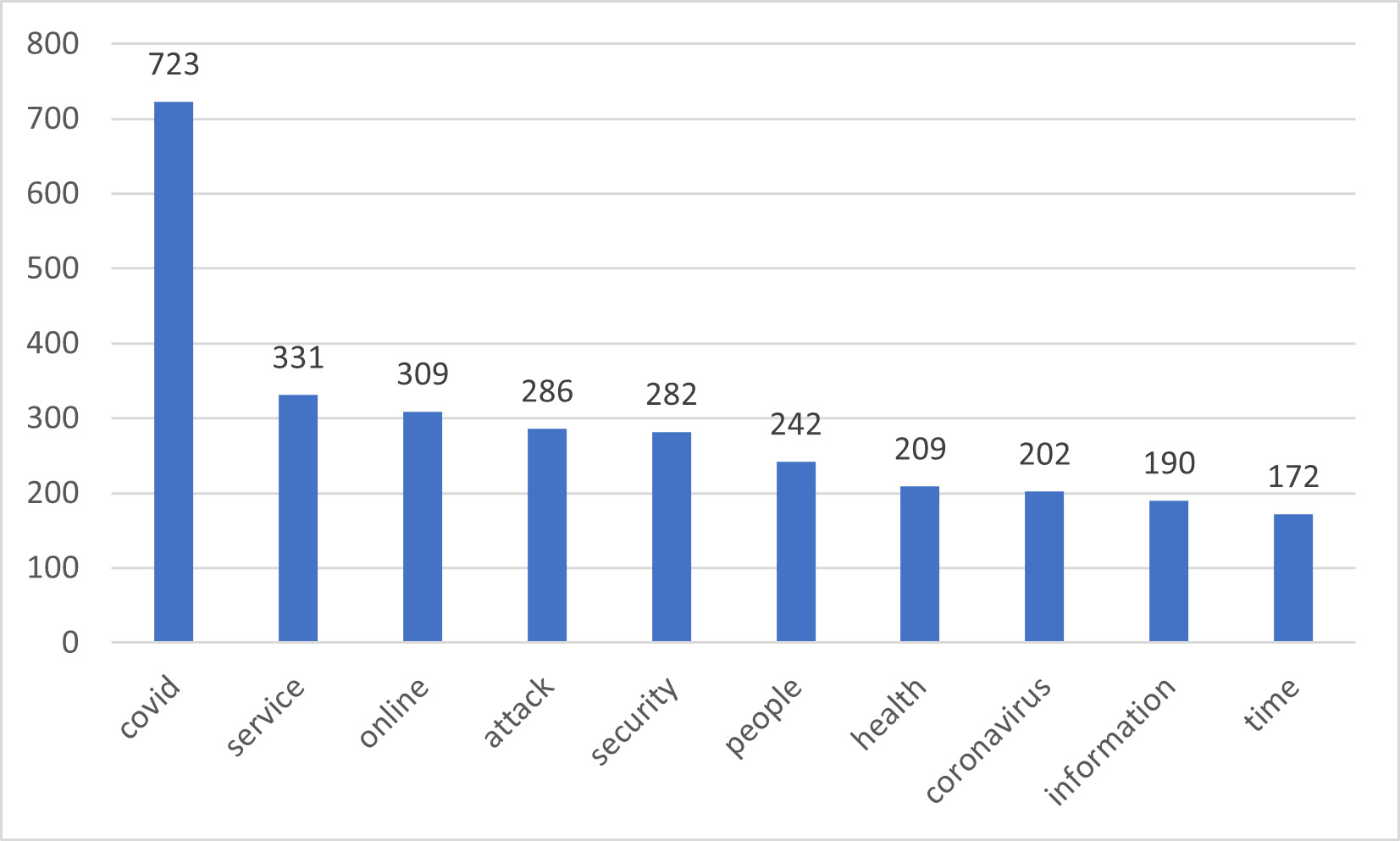}} \quad
        \subfigure[November]{\label{news-nov}\includegraphics[width=2.20in]{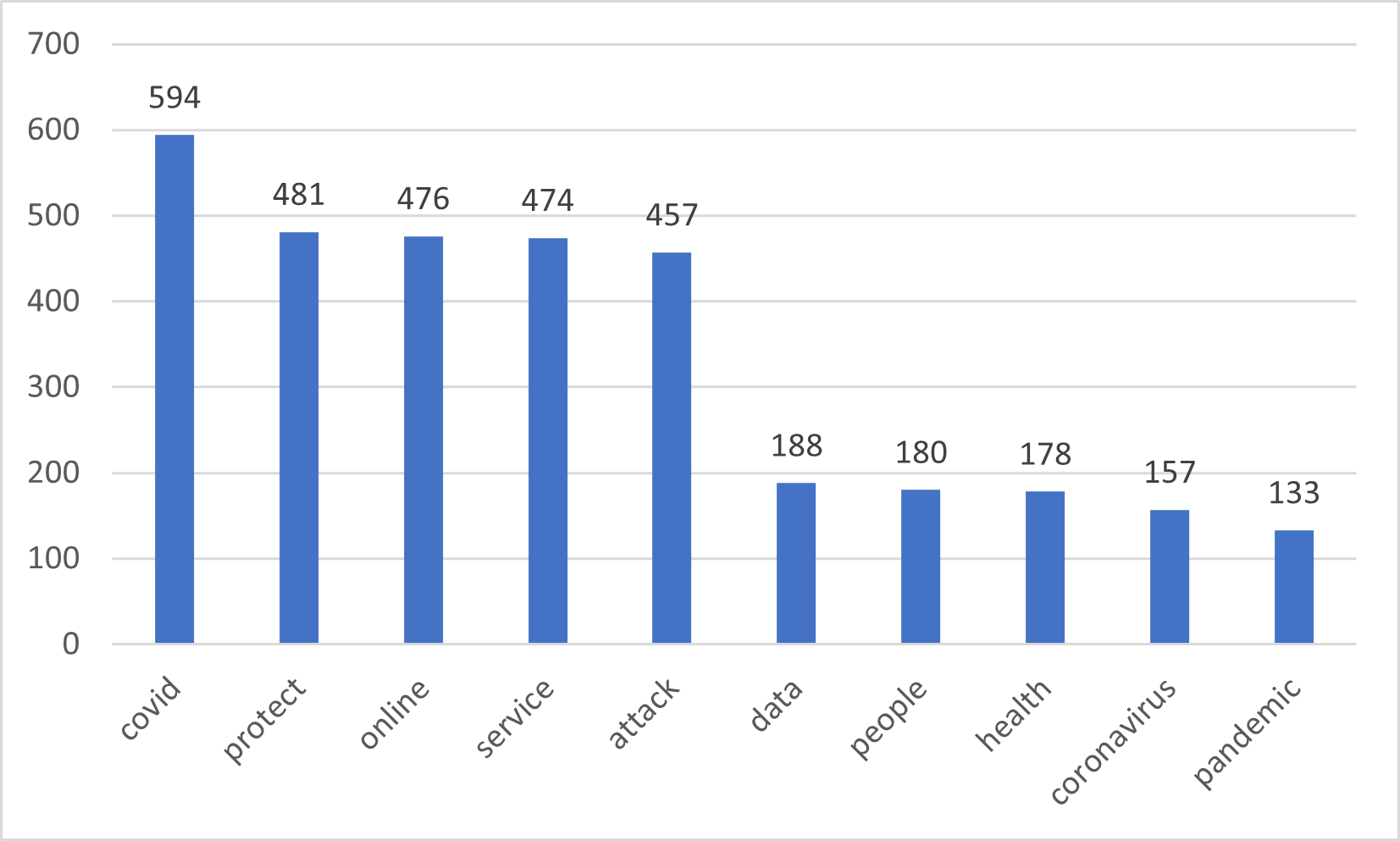}} \quad
        \caption{CoAID News Articles: Frequency of top 10 words.}
        \label{news-top10}
    \end{center}
\end{figure*}

\begin{figure*}[htb!]
    \begin{center}
        \subfigure[May]{\label{news-wordcloud-may}\includegraphics[width=2.20in]{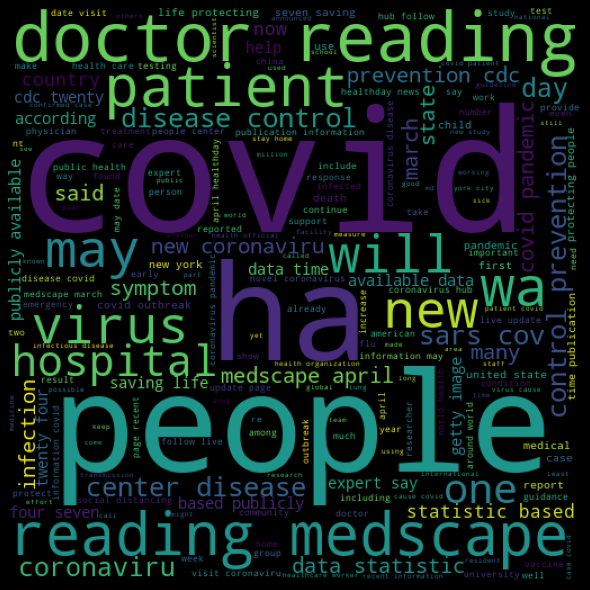}} \quad	
        \subfigure[July]{\label{news-wordcloud-july}\includegraphics[width=2.20in]{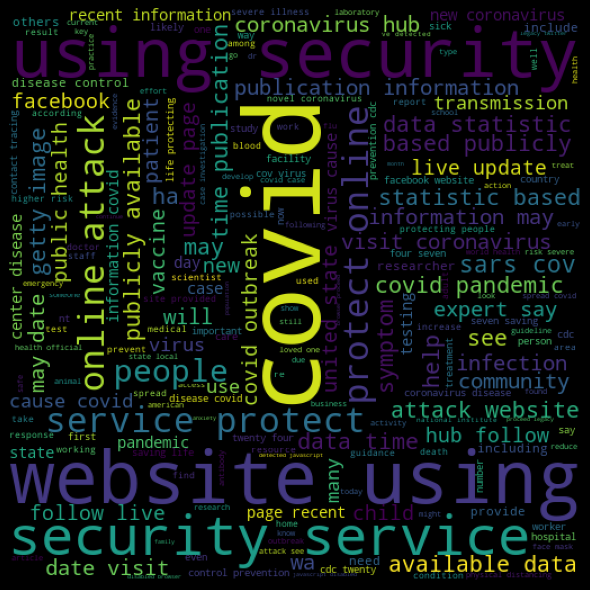}} \\	
        \subfigure[September]{\label{news-wordcloud-sept}\includegraphics[width=2.20in]{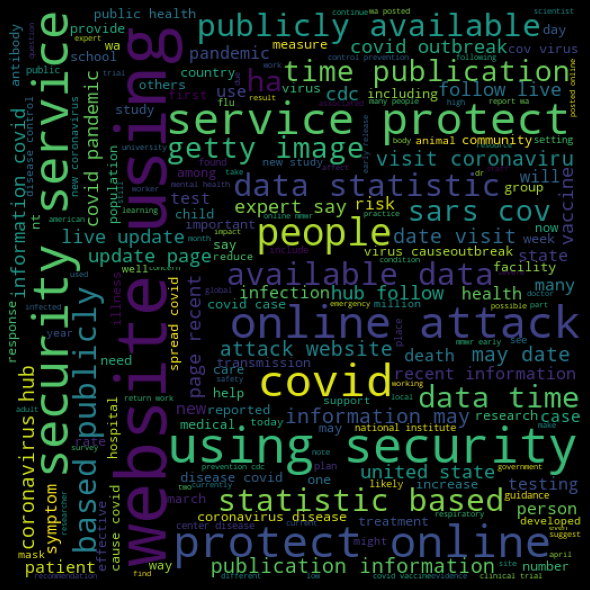}} \quad
        \subfigure[November]{\label{news-wordcloud-nov}\includegraphics[width=2.20in]{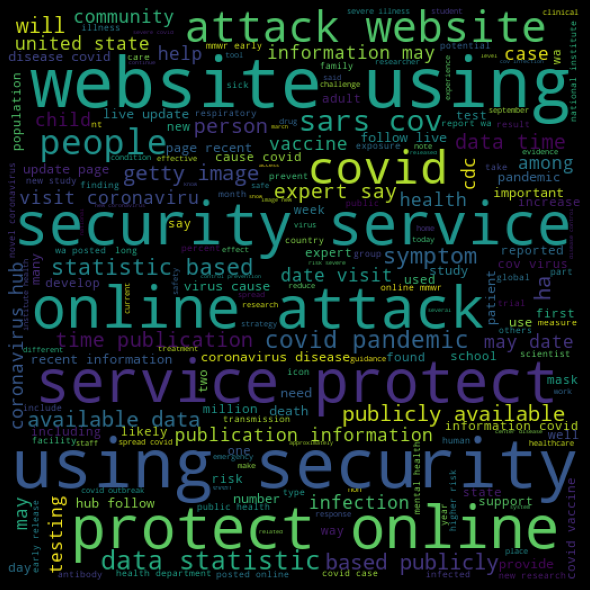}} \quad
        \caption{CoAID News Articles: Word Clouds for the four datasets.}
        \label{news-wordcloud}
    \end{center}
\end{figure*}

\paragraph{CoAID Sample Distribution.}

Figures \ref{fig:coaid-tweet} and \ref{fig:coaid-news} show the distributions of tweets and news articles per category, respectively. We see that the datasets contain significantly more true information than misinformation. Thus, the CoAID data is \textbf{heavily skewed}. For the tweets dataset, the sizes of May and July data are larger than that of September and November. Also, the number of misinformation tweets during September and November are negligibly smaller, making it challenging to use these as test datasets.

\paragraph{CMU Sample Distribution.}

The CMU-MisCov19 or CMU short-length text dataset misinformation categories are fine-grained comprising of 17 classes. It consists of 4,573 annotated tweets from 3,629 users with an average of 1.24 tweets per user. Figure \ref{fig:cmu} shows the \textbf{heavily skewed} distribution of 4,292 tweets for all categories that were extracted after some pre-processing. Class 7 (calling out or correction) has the most tweets, while class 2 (true treatment) has 0 tweets. 


\begin{table*}[!htb]
\begin{center}
\scalebox{1.0}{
\begin{tabular}{|p{2.0cm}|p{2.0cm}|p{1.4cm}|p{2.4cm}|p{2.2cm}|p{1.7cm}|p{1.6cm}|}  
\hline
     					& Embedding Dimension 		& \# Hidden Layers 			& Batch Size 		& Optimizer 		& Learning Rate \\ \hline
BERT         			&  768                   			&   12              			   	& 32           		& AdamW          	& 2e-5              \\ \hline
ELMo         			& 1024                    			&   5               			    	& 128           		& Adam          		& 0.001             \\ \hline
RoBERTa      			& 768                    			&   12             				&  16          		& AdamW           	& 2e-6             \\ \hline
ALBERT       			& 128                    			&   12             			  	& 6           			& Adam        		& 3e-5              \\ \hline
XLNet        			& 768                    			&  12                			  	&  32          		& AdamW          	& 2e-5               \\ \hline
Text GCN     			& 200                   			&   2              			 	& N/A          		& Adam           		& 0.02            \\ \hline
VGCN BERT    		        & N/A                    			&   18              			        & 16           		& Adam           		& 2e-5            \\ \hline
Word2Vec     			& 300                   			&   2              			        & 128           		& Adam          		& 0.001              \\ \hline
GloVe        			& 300                    			&   N/A              			        & N/A           		& AdaGrad       		& N/A               \\ \hline
\end{tabular}}
\caption{Experimental setting.}
\label{experimental_setting}
 \end{center}
\end{table*}

\subsubsection{CoAID News Articles: Study of Context Evolution}

Figure \ref{news-top10} shows context evolution in the news articles category via the evolution of the distribution of the top ten high-frequency terms. We see, similar to the tweet dataset, context changes over time in the news articles datasets. For example, The May and July datasets have only 4 common high-frequency words: covid, coronavirus, health, data. In the July dataset, we observe the emergence of three new high-frequency words attack, security, and protect, which indicates a change in context. The context in the September dataset seems to be similar to that of the July dataset. These two datasets have eight high-frequency words in common: covid, service, online, attack, security, health, information, coronavirus. The November dataset shares seven common words with the September dataset: covid, service, online, attack, people, health, coronavirus. However, we notice an increase in the frequency in some words such as protect and attack. Also, a new word pandemic is seen to emerge. We gather similar observations about the context evolution from the word clouds in Figure \ref{news-wordcloud}.

%

\section{Experimental Setting \& Training Statistics}

We provide the experimental setting for conducting our studies as well as the training statistics.

\paragraph{Experimental Setting.} Table \ref{experimental_setting} shows the experimental setting for the studies. We used the default learning rate and batch size for all experiments.


\begin{table*}[!htb]
\begin{center}
\scalebox{0.9}{
\begin{tabular}{|l|l|l|l|}
\hline
\textbf{Model}              & \textbf{\#Parameters} & \textbf{Dataset}                 & \textbf{Avg. Training Time} \\ \hline
\textbf{BERT PTM}           & 110M                  & CoAID News Articles (large data) & 1.25 mins                   \\ \hline
\textbf{BERT PTM}           & 110M                  & CoAID Tweets (large data)        & 1.66 hours                  \\ \hline
\textbf{BERT PTM}           & 110M                  & CoAID News Articles (small data) & 30 sec                         \\ \hline
\textbf{BERT PTM}           & 110M                  & CoAID Tweets (small data)        & 1.06 hours                  \\ \hline
\textbf{Word2Vec DSM + CNN} & \begin{tabular}[c]{@{}l@{}}Word2Vec: 6.7M\\ CNN: 160,301\end{tabular}   & CoAID News Articles (large data) & 3.48 mins  \\ \hline
\textbf{Word2Vec DSM + CNN} & \begin{tabular}[c]{@{}l@{}}Word2Vec: 155.9M\\ CNN: 160,301\end{tabular} & CoAID Tweets (large data)        & 1.17 hours \\ \hline
\textbf{Word2Vec DSM + CNN} & \begin{tabular}[c]{@{}l@{}}Word2Vec: 5.1M\\ CNN: 160,301\end{tabular}   & CoAID News Articles (small data) & 1.3 mins   \\ \hline
\textbf{Word2Vec DSM + CNN} & \begin{tabular}[c]{@{}l@{}}Word2Vec: 7.8M\\ CNN: 160,301\end{tabular}   & CoAID Tweets (small data)        & 46 mins    \\ \hline
\textbf{Word2Vec PTM + CNN} & 160,301               & CoAID News Articles (small data) & 2.05 mins                   \\ \hline
\textbf{Word2Vec PTM + CNN} & 160,301               & CoAID Tweets (small data)        & 2.2 hours                   \\ \hline
\textbf{GloVe DSM + CNN}    & \begin{tabular}[c]{@{}l@{}}GloVe: 4.6M\\ CNN: 160,301\end{tabular}      & CoAID News Articles (small data) & 1.33 mins  \\ \hline
\textbf{GloVe DSM + CNN}    & \begin{tabular}[c]{@{}l@{}}GloVe: 100.6M\\ CNN: 160,301\end{tabular}    & CoAID Tweets (small data)        & 45.25 mins \\ \hline
\textbf{GloVe PTM + CNN}    & 160,301               & CoAID News Articles (small data) & 2.08 mins                   \\ \hline
\textbf{GloVe PTM + CNN}    & 160,301               & CoAID Tweets (small data)        & 2.43 hours                  \\ \hline
\textbf{ELMo PTM + CNN}     & 160,301               & CoAID News Articles (small data) & 8.23 mins                   \\ \hline
\textbf{ELMo PTM + CNN}     & 160,301               & CoAID Tweets (small data)        & 8.63 hours                  \\ \hline
\textbf{Text GCN}           & N/A                   & CoAID Tweets (small data)        & 8.8 mins                    \\ \hline
\textbf{VGCN BERT}          & N/A                   & CoAID Tweets (small data)        & 33 hours                    \\ \hline
\textbf{Text GCN}           & N/A                   & CoAID News Articles (small data) & 7.32                        \\ \hline
\textbf{VGCN BERT}          & N/A                   & CoAID News Articles (small data) & 15.26 mins                  \\ \hline
\textbf{RoBERTa}            & 125M                  & CoAID News Articles (small data) & 5.39 mins                   \\ \hline
\textbf{ALBERT}             & 11M                   & CoAID News Articles (small data) & 6.8 hours                   \\ \hline
\textbf{XLNet}              & 110M                  & CoAID News Articles (small data) & 35 sec                      \\ \hline
\textbf{BERT PTM}           & 110M                  & CMU                              & 6 mins                      \\ \hline
\textbf{Word2Vec DSM + CNN} & \begin{tabular}[c]{@{}l@{}}Word2Vec: 8.7M\\ CNN: 160,301\end{tabular}   & CMU                              & 3 mins     \\ \hline
\textbf{Word2Vec PTM + CNN} & 160,301               & CMU                              & 30 sec                      \\ \hline
\textbf{GloVe DSM + CNN}    & \begin{tabular}[c]{@{}l@{}}GloVe: 8.6M\\ CNN: 160,301\end{tabular}      & CMU                              & 3.77 mins  \\ \hline
\textbf{GloVe PTM + CNN}    & 160,301               & CMU                              & 9.17 mins                   \\ \hline
\textbf{ELMo PTM + CNN}     & 160,301               & CMU                              & 1.05 mins                   \\ \hline
\end{tabular}}
\caption{Training Statistics - DSM: Domain-Specific Model, PTM: Pre-Trained Model}
\label{training_statistics}
\end{center}
\end{table*}

\paragraph{Training Statistics.}

Table \ref{training_statistics} shows the training statistics that include the number of parameters for each model, dataset, and the average training time. The inference time is not significant, thus not reported. All experiments were done on a Tesla V100 GPU, except the Text GCN and VGCN BERT based experiments, which were conducted using a CPU. For DSM and CNN based experiments, the CNN was trained for 5 epochs on the CoAID tweet data, 10 epochs on the CoAID news articles data, and 10 epochs on the CMU fine-grained data. The number of epochs was chosen based on the convergence behavior of the models.





\end{document}